%% file: journal-template.tex
\documentclass{myjournal}
\usepackage{mdframed}
\usepackage{amsmath}
\usepackage{tabularx}
\usepackage{array}
\usepackage{multirow}
\usepackage{booktabs}
\usepackage{makecell}
\usepackage{subfigure}
\usepackage{float}
\newcolumntype{L}[1]{>{\raggedright\arraybackslash}p{#1}}
\newcolumntype{C}[1]{>{\centering\arraybackslash}p{#1}}

\begin{document}

\title{Leveraging Non-Equilibrium ECRAM Dynamics for Short-Term Plasticity in Neuromorphic Circuits}



\author{Alex Currie$^1$,
Sean Borkholder$^1$,
Nithil Harris Manimaran$^2$,
Huayuan Han$^2$,
Cory Merkel$^3$,
Ke Xu$^2$*, and
Tejasvi Das$^1$*,
}

\affil{$^1$RAMLab, Department of Electrical and Microelectronic Engineering, Rochester Institute of Technology, Rochester NY, USA}

\affil{$^2$Iontronic and Nanoelectronics Lab, School of Physics and Astronomy, Rochester Institute of Technology, Rochester NY, USA}

\affil{$^3$Brain Lab, Department of Computer Engineering, Rochester Institute of Technology, Rochester NY, USA}

\affil{$^*$Author to whom any correspondence should be addressed.}

email: \textit{{tadeee@rit.edu, ke.xu@rit.edu}}

\vspace{4mm}

\begin{abstract}
Short-term plasticity (STP) is fundamental to temporal information processing in biological neural systems but remains difficult to realize efficiently in neuromorphic hardware. Memristive electrochemical random-access memory (ECRAM) devices naturally exhibit non-equilibrium ionic dynamics that produce transient conductance modulation; however, these behaviors are typically treated as undesirable variability or tolerated as side effects in memory-centric computing paradigms. In this work, we instead transform these volatile dynamics from a tolerated device artifact into a computational resource through a cross-layer device–circuit–system co-design framework. We introduce a delay-feedback leaky integrate-and-fire (LIF) neuron architecture co-designed with ECRAM synapses that exploits activity-dependent conductance modulation with negligible additional circuit overhead. The architecture integrates ECRAM-based synapses with a tunable delay-feedback spike-generation path, enabling transient device dynamics to directly modulate neuron excitability and synaptic efficacy. We used experimentally characterized ECRAM devices exhibiting transient conductance modulation ($\sim$1.5 k$\Omega$ per spike) to develop a compact behavioral model suitable for circuit-level simulation. Circuit simulations demonstrate two key STP behaviors -- synaptic facilitation and intrinsic excitability modulation -- while consuming ~2 pJ per spike, and the same device-driven mechanisms extend across multiple neuron topologies. Network-level analysis further demonstrates frequency-selective spike processing, allowing individual synapses to act as tunable temporal filters within spiking neural networks. This work demonstrates that non-equilibrium ECRAM dynamics can serve as a native hardware substrate for STP and temporal computation in neuromorphic circuits.

\end{abstract}

\section{Introduction}
As AI applications continue to grow rapidly in scope and deployment, the limitations of traditional von Neumann digital computing architectures have become apparent. As such, a primary focus of neuromorphic computing at the hardware scale is the development of area- and power-efficient compute-in-memory (CIM) circuit architectures. While approaches to neuromorphic circuit design have previously leveraged current-mode, charge-domain, and time-domain (TD) architectures, TD computing has gained increased attention due to its encoding of computational results entirely within signal delay \cite{Lou2024, Das2025}. Because no computational data is stored in the analog voltage and current characteristics of a TD-CIM signal, the outputs of such systems can be shaped as digital pulses and interpreted directly by sequential logic elements such as D Flip-Flops (DFFs). Conversely, current-mode and charge-domain architectures encode data within current flow and voltage levels, respectively. In order for such systems to effectively integrate into a traditional digital processing architecture, comparatively high area and power digital-to-analog converter (DAC) and analog-to-digital converter (ADC) circuits are required.

Several emerging technologies have been explored as memory storage elements for analog and mixed-signal TD-CIM circuitry. Generally, the criteria for use of such technologies is that they are able to hold a variety of nonvolatile conductance states depending on the history of biases applied to them, allowing convenient use as a variable R-C delay element in TD-NC circuits. Among emerging memristive technologies, two families stand out for additional volatile behavior under bias conditions that are too weak in amplitude or duration to trigger a nonvolatile state change: Memristors \cite{krestinskaya_neuromemristive_2020,hu_compact_2017} and Electrochemical RAM (ECRAM) \cite{tang_ecram_2018,um_ecram-based_2023}. In this work, we use the term ECRAM to refer to a memristive three-terminal ionically gated transistor platform that supports both nonvolatile electrochemical memory and volatile electric-double-layer dynamics, depending on biasing conditions. Both memristors and ECRAM change conductance values based on applied bias, but if the amplitude or duration of the applied bias is sufficiently low, a temporary increase in conductance is induced followed by a comparatively long decay back to the equilibrium conductance of the device's nonvolatile state. 

While such volatile dynamics are often dismissed as artifacts of physical implementation, we propose that they can be harnessed as a computational primitive in their own right. Prior works have touched upon how these volatile dynamics of memristive devices, including two-terminal memristors \cite{berdan_emulating_2016,ricci_tunable_2023} and three-terminal ECRAM-based \cite{rasetto_building_2023} neuromorphic architectures can be leveraged to emulate short-term plasticity (STP) observed in biological neurons, but ultimately remain abstractions of circuit-level behavior without any hardware-level implementation or direct exploitation of physical device dynamics. Circuit-level neuron models in existing literature typically treat synaptic weights as static or purely nonvolatile, with volatile behaviors either ignored or considered a byproduct of underlying device physics \cite{krestinskaya_neuromemristive_2020,indiveri_neuromorphic_2011}.

One key difference between memristors and ECRAM devices is the presence of a dedicated gate terminal in addition to a drain and source terminal in the latter, as opposed to the two terminals present in the former. The bias applied between the gate and source terminals of an ECRAM device determines the conductance between the drain and source terminals, allowing for separation of read and write operations and exceptionally low energy required to perform a write operation. In general, ECRAM also features a greater range of nonvolatile states, greater linearity of conductances across its range of states, and higher compatibility with analog systems than memristors \cite{lepri_-memory_2023,kang_two-_2021}. Crucially, this three-terminal topology enables selective access to the device’s volatile or nonvolatile behavior, which we exploit in this work to realize biologically inspired dynamic behavior.

\begin{figure}
\begin{mdframed}[linewidth=0.35pt,linecolor=black!35]
    \centering
    \includegraphics[width=0.9\linewidth]{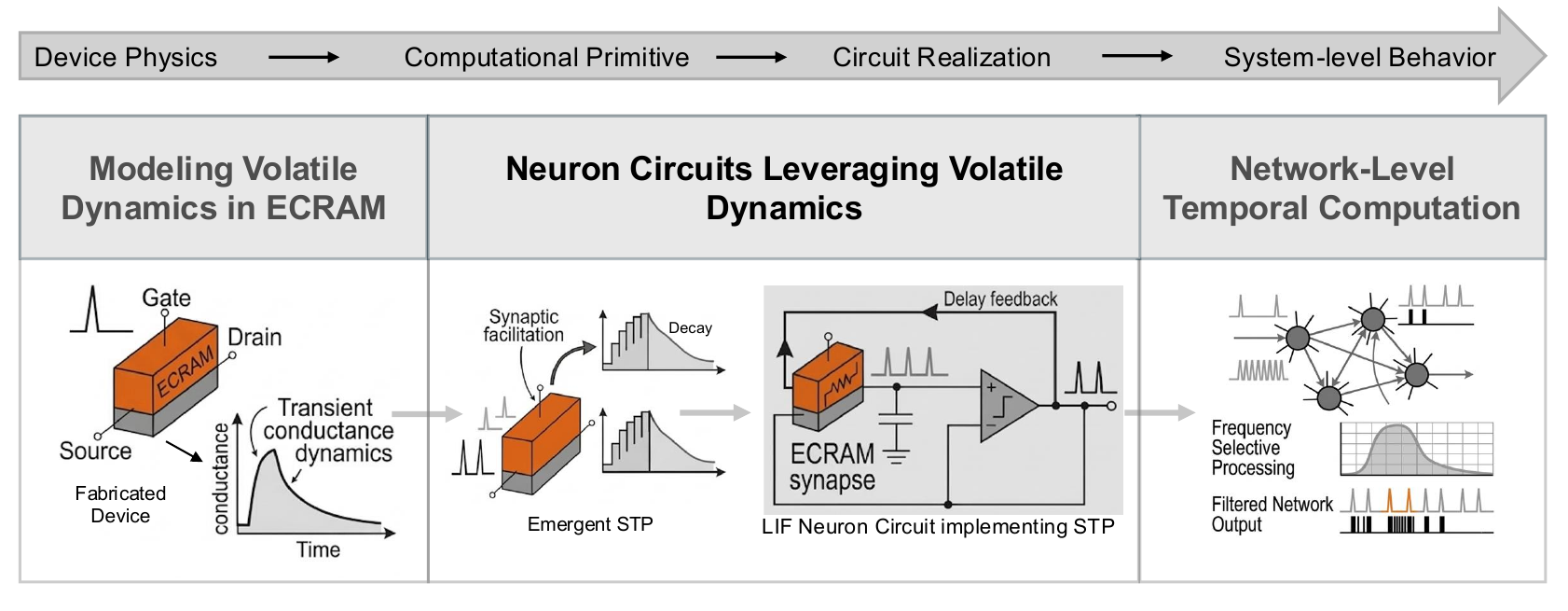}
    \vspace{-4mm}
    \caption{Proposed cross-layer framework linking volatile ECRAM device dynamics to neuromorphic computation: Volatile conductance dynamics in fabricated ECRAM devices are modeled and interpreted as a computational primitive corresponding to STP. These dynamics are exploited in neuron circuits to implement activity-dependent modulation within delay-feedback LIF architectures, enabling frequency-selective spike processing and temporal filtering at the network level.}
    \label{fig:blockdiag}
 \end{mdframed}
\vspace{-4mm}
\end{figure}

In this work, we intentionally exploit non-equilibrium memristive ECRAM dynamics not merely as a device artifact, but as a hardware mechanism for temporal computation. Through a cross-layer device–circuit–system co-design approach (illustrated in figure \ref{fig:blockdiag}), we show that volatile conductance dynamics in ionically gated devices can be harnessed through circuit topology to realize two complementary forms of STP directly in hardware: synaptic facilitation, in which recent presynaptic activity transiently strengthens an individual synapse, and intrinsic excitability modulation, in which postsynaptic spiking transiently increases the effective sensitivity of all incoming synapses. In this way, experimentally observed volatile dynamics become a circuit-level mechanism for activity-dependent modulation and, at the network level, a computational primitive for frequency-selective spike processing and temporal filtering. By selectively accessing volatile and nonvolatile behavior within the same ECRAM device, the proposed framework enables short-term temporal dynamics and long-term memory to coexist in a unified neuromorphic substrate.

\textbf{The primary contributions of this work are summarized as follows:}

\begin{enumerate}
\item \textbf{Device–circuit co-design leveraging volatile ECRAM dynamics}: We demonstrate that volatile non-equilibrium dynamics in ECRAM devices can be intentionally harnessed as a computational primitive for implementing STP in neuromorphic circuits.

\item \textbf{Delay-feedback LIF neuron architecture enabling dual STP mechanisms}: We introduce a LIF neuron circuit topology that explicitly exploits ECRAM volatile dynamics to realize two complementary forms of STP: local synaptic facilitation and global neuronal excitability modulation, with negligible additional circuit overhead – and demonstrate that the same mechanisms extend to a second LIF topology to illustrate architectural generality.

\item \textbf{Device-informed compact modeling framework}: Using experimentally characterized ECRAM devices, we develop a physics-informed compact behavioral model that captures transient conductance dynamics and enables accurate circuit-level simulation of STP behavior.

\item \textbf{Network-level temporal computation}: Through network-level analysis, we show that the proposed device--circuit mechanisms enable frequency-selective spike processing, allowing individual synapses to function as tunable temporal filters and enabling richer temporal computation in neuromorphic systems.
\end{enumerate}

The remainder of this paper is organized as follows. Section \ref{section:background} provides biological context and related work. Section \ref{section:ECRAM} describes the fabrication and modeling of the ECRAM devices used in this work. Section \ref{section:LIF} presents the proposed LIF neuron circuit architectures. Section \ref{section:results} presents device and circuit-level results with analyses. Section \ref{section:network} analyzes the network-level computational implications of the proposed mechanisms, followed by conclusions.

\section{Background and Motivation}\label{section:background}

Spiking neural network (SNN) architectures are a time-domain neuromorphic computing paradigm that have seen increased attention for their rich spatio-temporal dynamics in recent years \cite{M_Ferreira2025-ka,Zendrikov2023-ug}. SNNs are defined by networks of unit processor cells modeled after biological neurons, with information transmitted as a series of uniform spiking signals. Each spike is identical in shape, amplitude, and duration and, as is characteristic of TD-NC systems, information is encoded in the timing and relative delay of these spikes. This method of data encoding mimics the biological phenomenon of Action Potentials (APs), in which neurons produce a burst of potential that is transmitted to downstream cells through synaptic connections. 

Biological neurons operate by maintaining a resting electrical potential relative to their surrounding environment. As synaptic activity accumulates, the membrane potential of the neuron integrates the APs over time until it reaches a threshold, triggering an action potential at the neuron’s output through a cascade of cellular mechanisms. The Leaky Integrate and Fire (LIF) model captures key features of biological neurons, integrating spike inputs over time and firing when a threshold is exceeded, returning to its passive state if no spikes are received for an extended time  \cite{roy_towards_2019,el_ferdaoussi_efficiency_2023}. A typical SNN consists of thousands of LIF neuron elements connected to each other through weighted synaptic channels. Due to the scale of such systems, small and energy-efficient circuit implementations are essential for viable hardware deployment.

While the LIF neuron model captures first-order dynamics of biological spiking, biological neurons exhibit a range of additional functions that contribute to their computational power and efficient information transfer \cite{drukarch_thinking_2023}. Prior works have investigated modeling behaviors such as refractory period control, spike frequency adaptation, spike timing dependent plasticity, and short-term plasticity (STP) to enhance the computational ability of LIF neurons \cite{Lehmann_LIF_2022,deb_low-energy_2024,jooq_high-performance_2023,Joo_stdp_2022,kamal_ultra-low_2023,khacef_spike-based_2023}. While such functions add significant computational richness to the LIF model, they often require significant additional area and power overhead to implement at the hardware level. In this work, we focus on two biologically grounded STP mechanisms, synaptic facilitation and intrinsic excitability modulation, and show how these can be implemented efficiently using the volatile dynamics of ECRAM devices.

Synaptic facilitation is a type of STP that occurs at the synapse of biological neurons. As activity in the presynaptic neuron increases, synapses spatially near the activity will be facilitated and provide a greater relative contribution towards the generation of a postsynaptic AP. This synapse-specific effect reinforces frequently used pathways, enhancing temporal feature selectivity and acting as a high-pass filter for activity \cite{Jackman2017-nd}. 

Intrinsic excitability modulation is another form of STP enabled through the mechanism of neuronal facilitation in which the tendency of a neuron to generate an AP is modulated based on its short-term activity. Rather than affecting specific synaptic pathways, intrinsic excitability modulation increases a neuron’s overall responsiveness based on recent activity, supporting short-term memory and ensemble formation. In biological systems, intrinsic excitability modulation arises during an initial period of neuronal activity, increasing the firing likelihood of neurons that have recently spiked. This potentiation persists for a period of time after the initial activity is observed, allowing for short-term memory functions at the neuron scale and playing a critical role in long-term learning behaviors. Excitability modulation also aids in the formation of neuron ensembles, clusters of neurons that tend to exhibit increased activity as a group after receiving high levels of stimulation \cite{Hansel2024-nn}. Together, synaptic facilitation and intrinsic excitability modulation provide pathways to implement advanced computational behaviors at both the single neuron and network scales, making them key targets for implementation in SNN systems.

Efforts have been made to leverage the benefits of STP in SNNs, yet prior works have implemented solutions at the network scale with LIFs defined by abstract behavior rather than physical circuit implementation. SNNs leveraging STP have shown benefits including increased noise immunity and computational accuracy \cite{Leng2018-dt,Mejias2012-vt}, but often incur significant computational overhead when implemented in digital processing architectures. Limited work has been done to implement STP at the circuit-level through the use of synaptic structures, with switched capacitor and memristive synapses demonstrating efficacy in emulating STP behavior \cite{Noack_sc_synapse_2012,Mannan_2021}. However, such architectures suffer from significant area overhead and high static power consumption, respectively. Due to these costs, their scalability is severely limited, necessitating low-cost innovations to enable STP at the circuit-level.  

The volatile dynamics of ECRAM devices naturally mimic STP behavior, creating an opportunity for native circuit-level implementation without additional complexity. When a spike is applied to the gate terminal of an ECRAM synapse, the device conductance temporarily increases due to ionic redistribution within the channel and subsequently relaxes back to its equilibrium state over time. This transient conductance modulation effectively produces synaptic facilitation, where recent spiking activity temporarily increases synaptic efficacy. Repeated spike activity further elevates the average conductance level, enabling activity-dependent modulation of neuronal excitability that reflects recent input history. 

Because these dynamics arise from the intrinsic non-equilibrium behavior of the device rather than additional circuit elements, STP mechanisms can be realized with negligible hardware overhead. Importantly, these volatile dynamics coexist with the nonvolatile conductance states used for synaptic weight storage, allowing ECRAM devices to simultaneously support long-term memory and short-term temporal processing within the same physical synapse. 

\section{Physics-Based Modeling of ECRAM Devices}\label{section:ECRAM}

\subsection{Device Operation and Memory Mechanisms}

The devices considered in this work are ECRAM devices implemented using a three-terminal ionically gated transistor platform. In contrast to two-terminal memristive devices, the channel conductance is controlled by the gate-to-source voltage, while the drain-to-source bias is used only for readout. This separation of programming and sensing enables access to both transient and persistent conductance modulation within the same device while providing a natural mechanism for implementing short-term dynamics and long-term memory in neuromorphic circuits. The observed conductance modulation arises from distinct physical mechanisms depending on the applied gate bias. At low programming amplitudes or short pulse durations, conductance changes are dominated by electric double-layer (EDL) formation at the electrolyte-channel interface, producing a transient conductance increase that relaxes after removal of the gate bias \cite{mei_ionic_2024,xu_ionically_2025,xu_electric-double-layer-gated_2020}. At higher amplitudes or longer pulse durations, electrochemical doping induces persistent, nonvolatile changes in channel conductance \cite{talin_electrochemical_2025,talin_ecram_2023}. Together, these mechanisms enable combined short-term dynamics and long-term memory within a single device.

\subsection{Device Fabrication, Characterization, and Parameter Extraction}

\begin{figure}

    \begin{mdframed}[linewidth=0.35pt,linecolor=black!35]

    \centering
    \begin{minipage}{0.45\textwidth}
        \centering
        \includegraphics[width=0.75\textwidth]{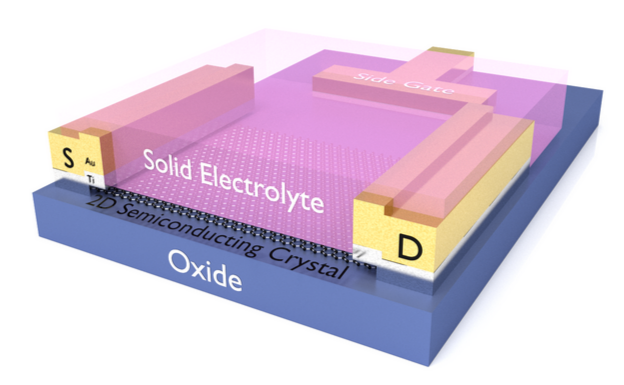}
        (a)
    \end{minipage}
    \begin{minipage}{0.45\textwidth}
        \centering
        \includegraphics[width=0.8\textwidth]{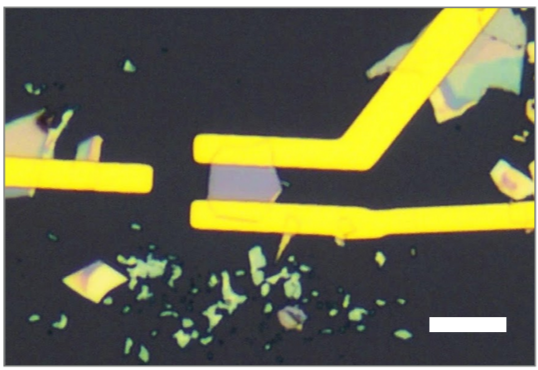}
        
        (b)
    \end{minipage}
    \begin{minipage}{0.45\textwidth}
        \centering
        \includegraphics[width=0.8\textwidth]{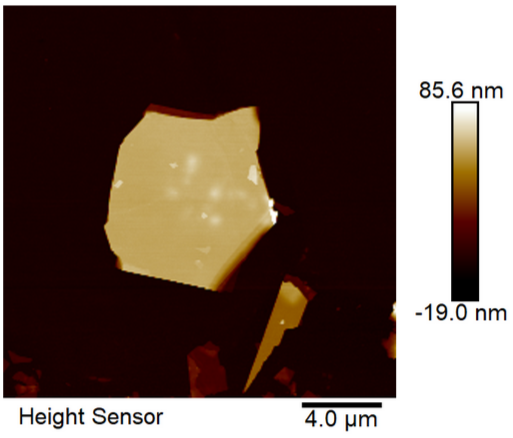}
        
        (c)
    \end{minipage}
    \begin{minipage}{0.45\textwidth}
        \centering
        \includegraphics[width=\textwidth]{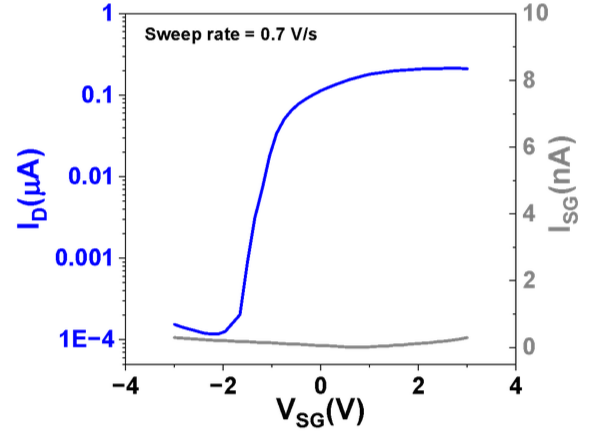}
        (d)
    \end{minipage}
    \caption{(a) Schematic of a side-gated ECRAM device. (b) Optical image of the fabricated device (scale bar = 8 $\mu m$). (c) Atomic Force Microscopy (AFM) scan of the $MoS_2$ channel before fabrication. (d) Ionically gated transfer characteristics of the fabricated device. V$_{DS}$ is held constant at 100 mV.}
    \label{fig:ecram_device}
    \end{mdframed}
\end{figure}

To develop and validate the physics-based ECRAM model used in this work, ECRAM devices were fabricated and parameterized using ionically gated transistors based on two-dimensional MoS$_2$ channels and a PEO:LiClO$_4$ solid polymer electrolyte. Devices were fabricated using standard planar microfabrication techniques, and representative characteristics have been reported previously \cite{xu_pulse_2018,han_energy_2025}. Figure \ref{fig:ecram_device} shows a schematic of a side-gated ECRAM and an optical image of a fabricated MoS$_2$-based device, along with channel surface characterization and ionically gated transfer characteristics. While the devices studied here operate primarily on millisecond timescales, prior work has shown that ionically gated transistors can achieve microsecond-scale dynamics through the use of high-mobility solid ion conductors \cite{nishioka_two_2025,takayanagi_ultrafast-switching_2023,cho_tuning_2024}, indicating that the modeling framework employed here is not intrinsically limited to a single timescale. Importantly, the circuit behaviors demonstrated in this work depend primarily on the relative timescales of transient conductance dynamics rather than the absolute device speed, allowing the same circuit principles to operate across a wide range of ionic transport regimes.

Time-resolved electrical measurements were performed on the fabricated devices to characterize both volatile and nonvolatile device behavior. Nonvolatile conductance states were established using sequences of potentiation and depression pulses spanning 128 experimentally demonstrated states \cite{han_energy_2025}. Volatile dynamics were characterized by applying single 2.2-s gate voltage pulses spanning 0.1–2.5 V and measuring the resulting transient drain current response at a fixed drain-source voltage of 0.1 V, capturing both conductance rise and relaxation following pulse termination. 

These measurements were used to extract ionic relaxation timescales and drift–diffusion–dependent parameters governing ion motion and electric double-layer formation. Using these experimentally extracted parameters, a physics-based compact device model was developed to capture both volatile and nonvolatile conductance dynamics. The extracted ionic dynamics were coupled to the Enz–Krummenacher–Vittoz (EKV) transistor framework to map ion-induced electrostatic gating onto channel current, following the general modeling approach of Sun et al \cite{sun_dynamic_nodate}. The resulting device model was first created in Python, verified against experimental device measurements, and subsequently translated to Verilog-A for direct integration into the Cadence circuit simulation tools used for this work.  The accuracy of the model is demonstrated in Section \ref{section:results} through direct comparison between simulated and experimentally measured data.


Although early ionically gated transistors relied on thick liquid or gel electrolytes, recent developments in ionically gated transistor technologies have demonstrated solid-state electrolytes with nanoscale dimensions and potential compatibility with CMOS back-end-of-line processing \cite{zhang_electrolyte-gated_2024,min_cmos-compatible_2020,lee_understanding_2023,liu_bionic_2023,waser_redox-based_2009,paletti_two-dimensional_2019,awate_impact_2023,liang_molecularly_2019,xu_monolayer_2017}. These advances indicate a viable pathway toward scalable integration; however, to maintain the highest possible level of accuracy towards our proposed circuit-level solutions, we primarily focus on our experimentally validated device model for simulations in this work and do not directly extrapolate to other electrolyte compositions. 

While the specific parameters used in this work correspond to MoS$_2$ ECRAM devices with a PEO:LiClO$_4$ electrolyte, the modeling framework is general and can be adapted to other ionically gated transistor platforms by updating the extracted ionic transport parameters. The compact model developed here is used directly in all circuit- and network-level simulations in this work, enabling circuit architectures that exploit the transient conductance dynamics of ECRAM devices.

\section{LIF Circuit Framework Leveraging Volatile ECRAM Dynamics}\label{section:LIF}

To demonstrate the application of ECRAM volatile ion dynamics to hardware implementation of STP, we introduce two LIF neuron circuits using ECRAM-based synapses. While basic LIF functionality is maintained by treating ECRAM devices as purely nonvolatile elements, STP can be achieved by leveraging the ECRAM gate terminal to couple existing circuit signals into the ionic state, allowing internal activity to produce non-equilibrium ion dynamics and modulate circuit behavior. This approach requires no additional circuit elements and thus incurs negligible additional area overhead. We introduce a novel LIF circuit topology optimized for integration with ECRAM-based synapses. We further demonstrate that the proposed techniques can also be readily extended to other LIF neuron circuit architectures with minimal modification. In this framework, temporary conductance modulation arises when activity-based signals are applied to the ECRAM gate, producing non-equilibrium ionic dynamics that modify synaptic current and enable short-term plasticity behaviors.

\subsection{ECRAM-based synapse}
\begin{figure}

    \begin{mdframed}[linewidth=0.35pt,linecolor=black!35]

    \begin{minipage}{0.5\textwidth}
    \hspace*{-2cm}
    \centering
    \includegraphics[width=0.5\textwidth,
    trim = 20 -10 20 20, clip]{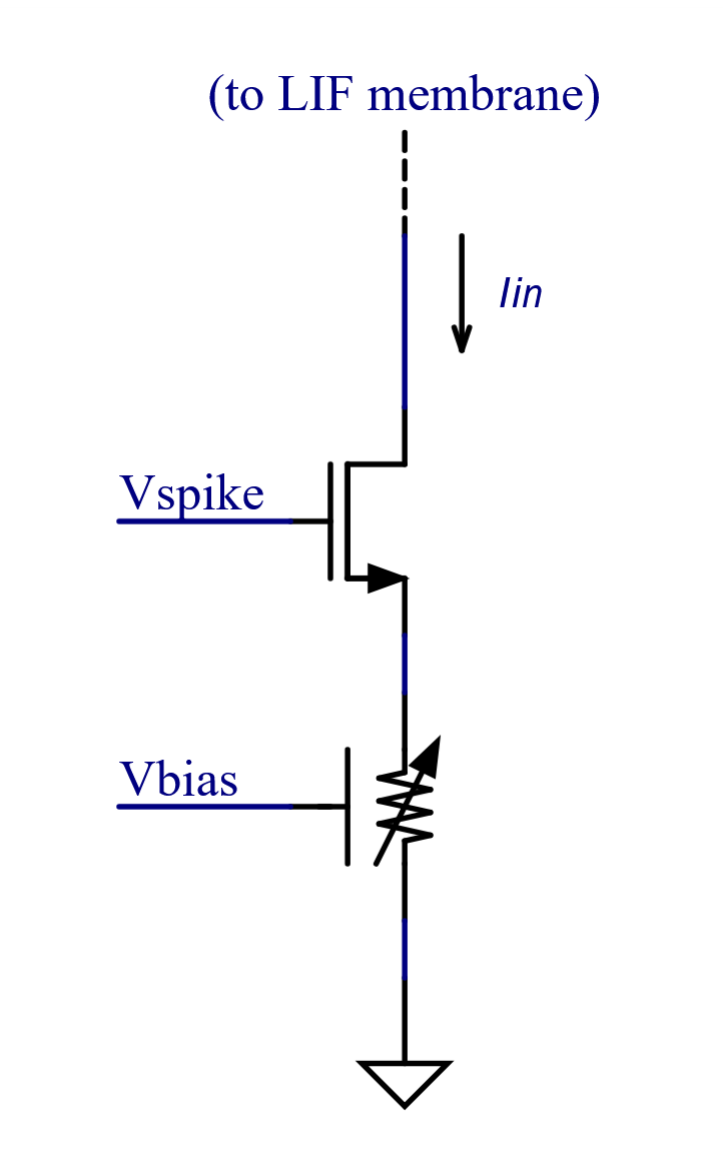}
    \hspace*{-3cm}
    (a)
    \end{minipage}
    \begin{minipage}{0.5\textwidth}
    \hspace*{-1.75cm}
        \includegraphics[width=1.2\textwidth, 
        trim= 20 0 190 20, clip]{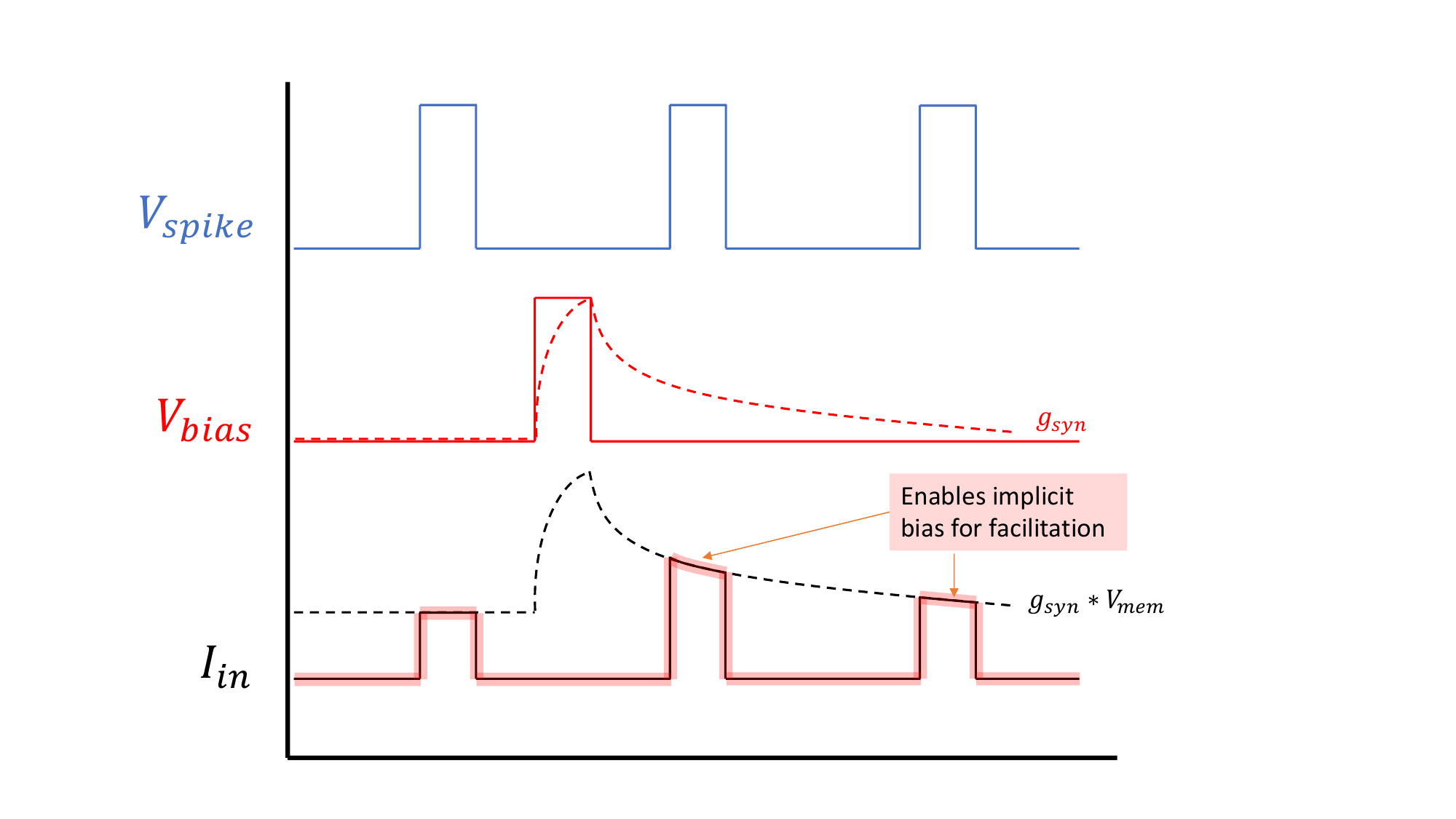}
        \centering
        (b)
    \end{minipage}
    \caption{(a) Proposed 1T1E synapse structure. A HVT NMOS transistor acts as the primary synaptic input, only allowing current to flow when a voltage pulse is applied to its gate. An ECRAM device acts as a synaptic weight, restricting current flow based on its nonvolatile conductance state. Current can be subsequently integrated by a connected LIF circuit structure. (b) By applying a voltage pulse to the ECRAM gate in conjunction with standard spiking inputs, the amplitude of resultant current pulses can be temporarily increased over an extended time period, emulating STP behaviors of biological neurons.}
    \label{fig:synapse-1T1E}
    \end{mdframed}
\end{figure}

A common application of synaptic devices in neuromorphic architectures is the 1T1R structure, in which a series connection between a resistive memory element and a transistor acting as a switch emulates a biological synapse \cite{krestinskaya_neuromemristive_2020, lepri_-memory_2023}. The transistor gate acts as a synaptic input by presenting an effective open circuit when no bias is applied and allowing current flow when a voltage pulse is presented. For LIF circuit designs, this behavior can be leveraged to provide a high-impedance input for spiking voltage signals and integrating the resulting current pulses using capacitive elements. Our implementation similarly utilizes an ECRAM device in series with an NMOS device to form a 1T1E synaptic structure (figure \ref{fig:synapse-1T1E} (a)), where the ECRAM provides the programmable synaptic weight while its gate terminal enables access to volatile ionic dynamics. To minimize leakage currents over long timescales, we have chosen a high threshold voltage (HVT) device provided in the GlobalFoundries 55 nm PDK for the switching device and maintained minimum sizing. The low component count and parasitic capacitance of the ECRAM allow for high local scalability, with an arbitrary number of synaptic inputs able to be integrated into a single LIF circuit. This enables flexibility at the network scale, as the number of possible interconnections between LIF primitives is not limited by inherent circuit behaviors.

For normal operation, inputs drive the gate of the transistor and the ECRAM acts as a fixed-conductance device. Assuming negligible on-state resistance of the FET device relative to the ECRAM device, the current flow through the synapse is determined by the product of the membrane potential and nonvolatile ECRAM conductance. By additionally routing pulsed signals to the ECRAM gate, transient ionic redistribution can be induced, enabling non-equilibrium ion dynamics that can be leveraged for advanced computational behavior (figure \ref{fig:synapse-1T1E} (b)). 

Two distinct short-term plasticity behaviors can be implemented using this mechanism. For instance, by connecting the ECRAM gate to its corresponding transistor gate input, conductance is temporarily increased when local synaptic activity occurs, emulating the synaptic facilitation seen in biological neurons. If the ECRAM gates of all synapses in a given LIF are connected to that neuron's output, then every synapse is temporarily facilitated when an output spike is generated. While the actual firing threshold of the LIF remains static, this transient increase in synaptic conductance effectively biases the neuron toward greater input sensitivity, producing a lower effective firing threshold and thereby emulating intrinsic excitability modulation. This synaptic structure forms the input stage of the LIF neuron architectures described in the following sections.

\subsection{Proposed LIF Delay-Feedback Neuron Architecture}
\begin{figure}
    \centering
    \begin{mdframed}[linewidth=0.35pt,linecolor=black!35]
    \centering
    \includegraphics[width=\textwidth]{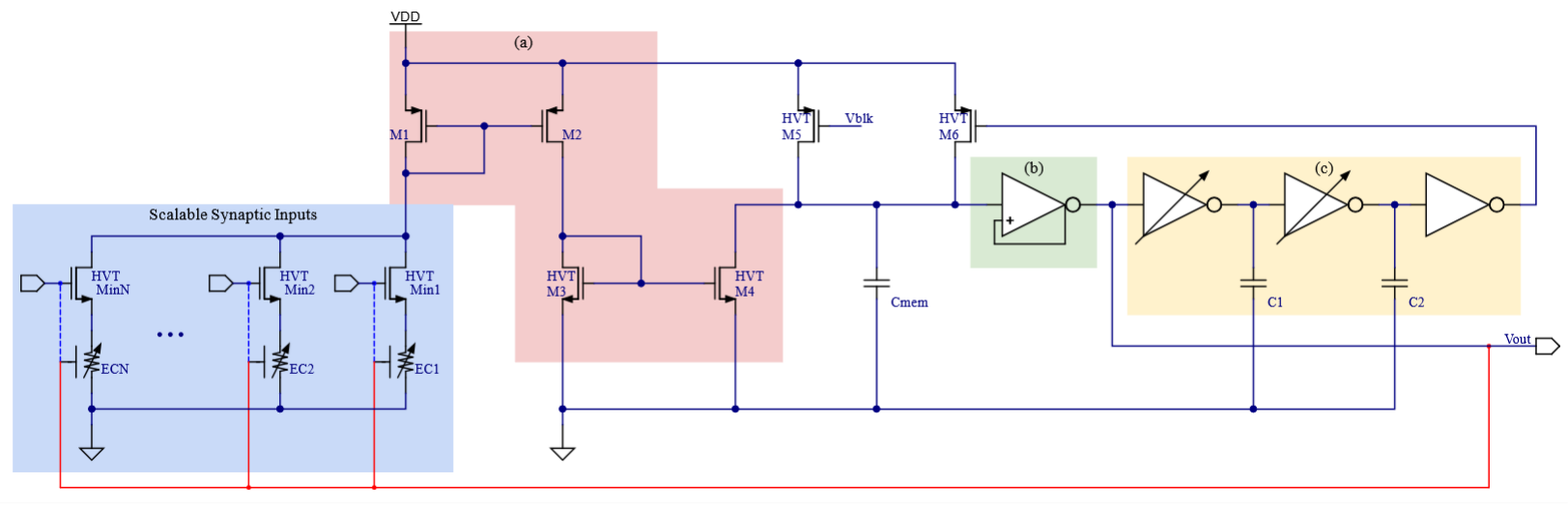}
    \caption{Proposed delay-feedback LIF neuron architecture co-designed for operation with volatile ECRAM synapses. Input current mirroring (a) integrates synaptic currents onto the membrane capacitor $C_{mem}$, while a positive-feedback firing stage (b) generates robust output spikes and suppresses near-threshold shoot-through currents. A delay-feedback path formed by current-starved inverters and capacitive loading (c) controls spike width and refractory behavior, enabling neuron dynamics compatible with the volatile timescales of ECRAM devices. STP behaviors are implemented by routing activity-dependent signals to the ECRAM gates.}
    \label{fig:novel_LIF}
    \end{mdframed}
\end{figure}

To demonstrate hardware implementation of STP using volatile ECRAM dynamics, we propose a delay-feedback LIF neuron architecture, shown in figure \ref{fig:novel_LIF}. The neuron employs ECRAM-based synapses and a precharged membrane node that integrates synaptic current while exhibiting controlled leakage dynamics. Membrane integration and leak behavior establish the core LIF dynamics, while spike generation and refractory timing are governed by a delay-feedback path composed of current-starved inverters. By biasing the current-starving transistors in deep subthreshold, the feedback path introduces a tunable delay that enables millisecond-scale spike timing compatible with the ion dynamics of the fabricated ECRAM devices. This architecture therefore preserves the baseline behaviors of a standard LIF neuron (as discussed in Section \ref{section:background}) while enabling STP through activity-dependent modulation of the ECRAM gate potential in each synapse.

STP behaviors can be implemented by routing activity-dependent signals to the ECRAM gates. Connecting each ECRAM gate to its corresponding NMOS gate (blue dashed lines in figure \ref{fig:novel_LIF}) enables synaptic facilitation through activity-dependent transient modulation of the synaptic conductance. Alternatively, connecting the ECRAM gates to the neuron output (red solid lines) enables intrinsic excitability modulation, where output spikes temporarily increase the effective synaptic conductance across all inputs. Only one STP configuration is active at a time.

By leveraging the synapse structure shown in figure \ref{fig:synapse-1T1E}, incoming voltage pulses are converted to a current proportional to the ECRAM conductance state. Synaptic current is delivered to the membrane capacitor $C_{mem}$ using the current mirror structures of M1/M2 and M3/M4, where successive synaptic events incrementally discharge the precharged membrane node from its initial voltage of $VDD$. In this way, $C_{mem}$ integrates synaptic current over time, implementing the core membrane accumulation behavior of the LIF model. As the voltage drop on $C_{mem}$ is proportional to the product of the capacitance and the integral of the current, a potential issue arises in which millisecond-scale current pulses could completely discharge the capacitor regardless of the programmed ECRAM conductance, defeating the purpose of varied synaptic weights. To combat this non-ideality without drastically increasing capacitor size and thus circuit area, the current mirroring structure is scaled such that every input synaptic current pulse is divided by a significant amount before being delivered to $C_{mem}$. Because the mirrors are cascaded, the total current scaling factor is the product of the individual mirror ratios (M1–M2 and M3–M4). This allows the desired current attenuation to be achieved using smaller devices than a single large mirror stage. The use of an input current mirror also benefits circuit operation by isolating synaptic connections from the membrane node, allowing for an arbitrary number of synaptic connections to be made without impacting membrane capacitance.

The leaky integration function of the delay-feedback LIF neuron is enabled by the combination of $C_{mem}$ and M6. While the voltage across $C_{mem}$ ($V_{mem}$) represents the accumulated integral of incoming synaptic current, M6 acts in the subthreshold region to slowly restore $V_{mem}$ to $VDD$. The amount of leakage is externally controlled by the gate voltage $V_{blk}$. Only when the frequency of input synaptic activity is sufficient to overcome leakage will $V_{mem}$ be driven below the switching threshold voltage of a positive feedback inverter (see section \ref{subsubsection:pos_fb_inverter}), initiating a firing event as the inverter output transitions to high. The use of the positive feedback inverter to initiate an output AP ensures that the membrane voltage does not remain near the firing threshold for an extended period of time. As a result, shoot-through currents are mitigated and output pulses are strongly driven to the supply rails ($VDD$ and $VSS$).

As the output ($V_{out}$) is driven high, a feedback path composed of two current-starved inverters and a standard CMOS inverter turns on M7 after a controlled delay, restoring $V_{mem}$ to $VDD$ and resetting the membrane node. As a consequence, the positive feedback inverter changes state again and the output pulse ends. M7 remains on until the falling edge of the output pulse propagates through the inverter chain, continuing to pull $V_{mem}$ upward and temporarily suppressing the effects of incoming synaptic activity. This behavior emulates the refractory period seen in biological neurons after a firing event. The length of the output pulse and refractory period is determined by the amount of time it takes for the signal to travel through the inverter chain on the rising and falling edge of $V_{out}$, respectively, and can thus be tuned by modulating the amount of current starving in the inverter chain. Capacitors C1 and C2 were incorporated into the output of each current-starved inverter to further slow the signal, achieving a pulse time of 1 ms for ECRAM functionality. Overall, this novel topology presents a low transistor-count, scalable, and highly tunable solution for implementing the LIF neuron model at the circuit-level, while naturally supporting volatile behavior-enabled STP mechanisms.

\subsection{Spike Generation and Output Stage Circuits}

Ideally, an individual LIF neuron circuit accepts digital voltage pulses at its inputs and produces digital voltage pulses at its output. This both emulates biological action potentials, which are typically fixed in amplitude and duration at the local scale, and allows for arbitrary scalability and integration with digital systems. However, the internal operation of the LIF neuron relies on analog membrane dynamics, and therefore the circuitry within the neuron need not be constrained to digital operation. Converting a digital input pulse to an analog signal can be achieved through the use of weighted synapses such as those proposed in the previous section, but generating a digital output pulse based on analog circuit conditions without costly components such as traditional comparators can pose challenges. A CMOS inverter (figure \ref{fig:inverters} (a)) can act as a low-cost comparator with its threshold set midway between $VDD$ and $VSS$, but suffers from high shoot-through currents when $V_{in}$ is near the switching threshold. Implementing leakage in LIF neuron circuitry also adds complexity to ensuring predictable output behavior. When synaptic activity causes the membrane voltage to exceed the action potential threshold, leakage current can immediately pull the voltage back below the threshold, potentially causing incomplete or unstable output transitions. While these issues remain true for the two-transistor CMOS inverter, several compact inverter-based circuit variants can be employed to improve spike generation and output stability, as described in the following subsections.

\subsubsection{CMOS Schmitt Trigger Inverter}
A straightforward approach to eliminate unwanted toggling caused by membrane voltage fluctuation is to implement hysteresis using a CMOS Schmitt trigger inverter \cite{filanovsky_cmos_1994} (figure \ref{fig:inverters} (b)). The resulting double-threshold switching behavior ensures that once the input crosses the LIF firing threshold, the membrane voltage must fall significantly before the output toggles again. This prevents spurious switching due to membrane voltage fluctuations and allows the rest of the circuit sufficient time to generate a complete output spike before the neuron resets.

\subsubsection{Positive Feedback Inverter}\label{subsubsection:pos_fb_inverter}
A secondary solution to incomplete threshold crossing is the implementation of a positive feedback inverter (figure \ref{fig:inverters} (c)). This circuit actively resolves near-threshold input voltages by applying positive feedback \cite{Culurciello_FB_INV_2003}, pulling an input floating near mid-rail toward $VSS$ to ensure that the output voltage resolves to a strong $VDD$ while preventing prolonged shoot-through currents. When the input voltage is well above the switching threshold, M3 acts to pull the output to $VSS$. M3 is chosen to be an HVT device to ensure that on a falling edge at the input it turns off before M4–M6 such that the latter devices dominate during transient switching behavior. M3 is required due to the inability of the diode-connected transistor M5 to pull the output voltage fully to $VSS$, and thus maintains rail-to-rail output behavior. At near-threshold input voltage, both M1 and M2 operate in the saturation region and allow shoot-through currents to pass from $VDD$ to $VSS$. M4 and M5 mirror this current back to the input node, actively pulling the input downward in an attempt to force it to $VSS$, at which point only M1 remains active and nominally no current flows through the circuit. M6 ensures that this pull-down path is turned off when the input is near $VDD$ to further minimize leakage currents at steady state. It is chosen to be an LVT device such that it turns on before the M1/M2 switching threshold to enable the positive feedback action that rapidly resolves the output transition.

\subsubsection{Current Starved Inverter}
\begin{figure}
    \centering
    \begin{mdframed}[linewidth=0.35pt,linecolor=black!35]
    \centering
    \begin{minipage}{0.15\textwidth}
        \centering
        \includegraphics[width=\textwidth]{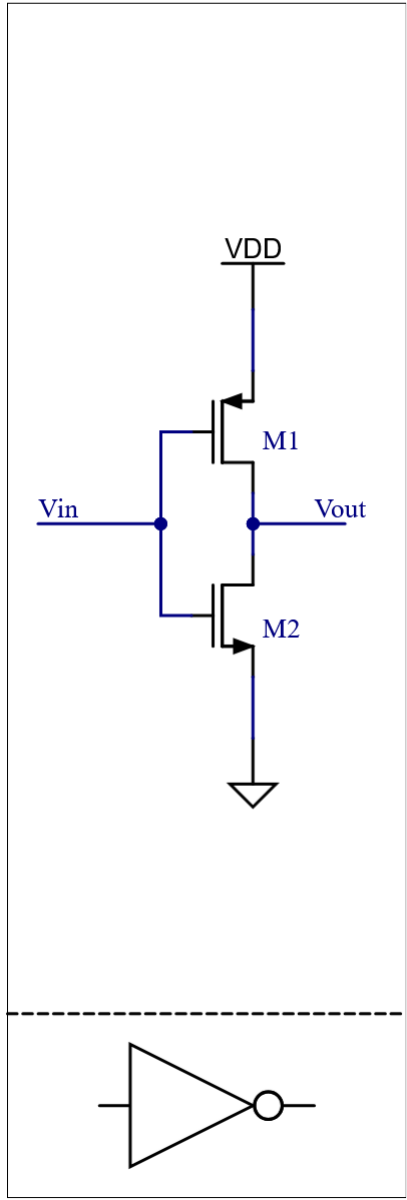}
        (a)
    \end{minipage}
    \begin{minipage}{0.265\textwidth}
        \centering
        \includegraphics[width=\textwidth]{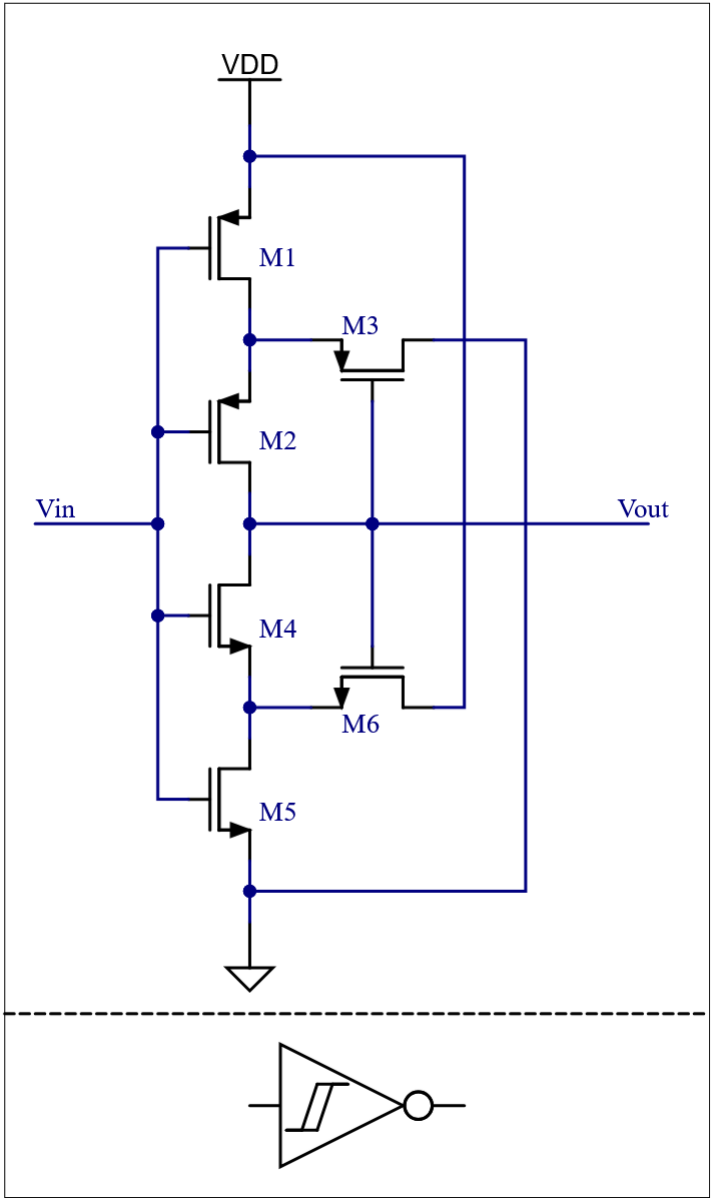}
        (b)
    \end{minipage}
    \begin{minipage}{0.265\textwidth}
        \centering
        \includegraphics[width=\textwidth]{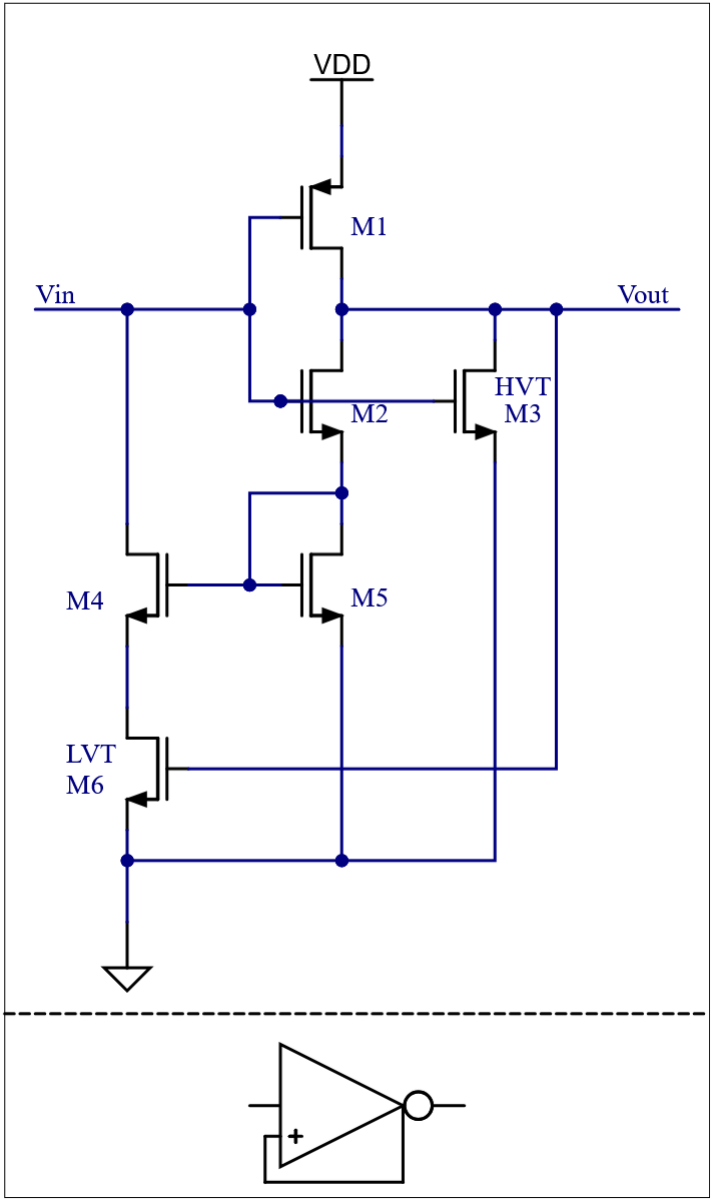}
        (c)
    \end{minipage}
    \begin{minipage}{0.15\textwidth}
        \centering
        \includegraphics[width=\textwidth]{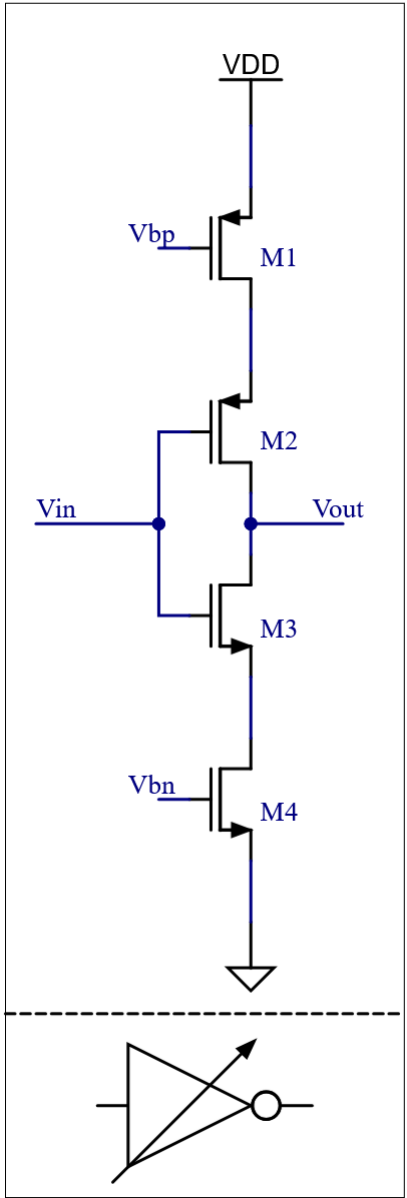}
        (d)
    \end{minipage}
    \caption{Inverter variant schematics and symbols used in our LIF designs: (a) standard CMOS inverter, (b) CMOS Schmitt trigger inverter, (c) positive feedback inverter and (d) current-starved inverter.}
    \label{fig:inverters}
    \end{mdframed}
\end{figure}

Lastly, by intentionally limiting the current an inverter can source or sink, its switching action can be significantly slowed, enabling controlled timing behavior within the neuron circuit. This can be achieved through current-starving, in which a transistor biased in the subthreshold region is added to the pull-up and pull-down paths of the inverter (figure \ref{fig:inverters} (d)). While current-starving alone may not solve the issue of near-threshold operation, it can be leveraged in conjunction with other inverter topologies to introduce controllable delays and precisely shape spike generation and reset behavior.

The proposed neuron architecture establishes a hardware framework for leveraging volatile ECRAM dynamics. When integrated within larger spiking networks, the activity-dependent modulation produced by these circuits naturally gives rise to frequency-selective spike processing, which we analyze at the network level in Section \ref{section:network}. The following sections examine the implications of this design for timing scalability and compatibility with other neuromorphic circuit implementations.

\section{Timescale and Throughput Scaling in ECRAM-Based LIF Circuits}

As presented, our novel LIF topology is designed to operate at the millisecond timescale of the demonstrated PEO:LiClO$_4$-based ECRAM synapses, ensuring that neuron dynamics remain compatible with the measured ionic behavior of the devices. However, recent works have demonstrated that ECRAM devices can exhibit volatile dynamics acting on the microsecond scale given electrolyte material choice and engineering \cite{nishioka_two_2025,takayanagi_ultrafast-switching_2023,cho_tuning_2024}. The integration of such ECRAM devices into the proposed circuit would require significantly less deliberate slowing of the neuron dynamics, while simultaneously enabling greater computational throughput. 

To demonstrate the benefits of faster-acting ECRAM devices for STP implementation, we have created a variant of our device model in which parameters have been adjusted to allow for shorter ion dynamics time constants. This allows us to reduce the pulse time of each AP to the microsecond scale, representing a potential speedup of $10^3\times$. The output spike width of the LIF can easily be matched to this faster regime by adapting current-starving bias voltages in the feedback path. At the microsecond timescale, capacitors C1 and C2 can be entirely removed, with current starving alone used to dictate output characteristics. 

Additionally, because input current pulses have a significantly shorter duration, the total amount of integrated charge per current spike is drastically diminished at shorter timescales, allowing for less aggressive current division in the input mirror and significantly reduced size of $C_{mem}$. These modifications naturally lead to more compact and scalable neuron circuits, which are desirable for large-scale neuromorphic network integration.
\section{Generalization of Volatile ECRAM Dynamics to other LIF Architectures}

Several recent works have presented circuit-level implementations of the LIF neuron model, with integration of input current pulses onto a capacitive element being a popular implementation of the core LIF behavior and voltage pulses representing input and output APs \cite{jooq_high-performance_2023,deb_low-energy_2024,aamir_highly_2016,cho_cmos_2025}. Because these architectures rely on weighted synaptic current integration, the ECRAM-based synapse proposed in this work can be readily incorporated into a wide variety of existing circuit implementations.

The LIF circuit presented in \cite{jooq_high-performance_2023}, hereafter referred to as the capacitive refractory topology, is particularly compact. It features a low overall transistor count and no complex sub-circuits such as amplifiers or comparators, making it a useful case study for demonstrating  how ECRAM-enabled STP can be incorporated into other scalable neuron circuits.

\begin{figure}
    \centering
    \begin{mdframed}[linewidth=0.35pt,linecolor=black!35]
   \centering
    \includegraphics[width=0.8\textwidth]{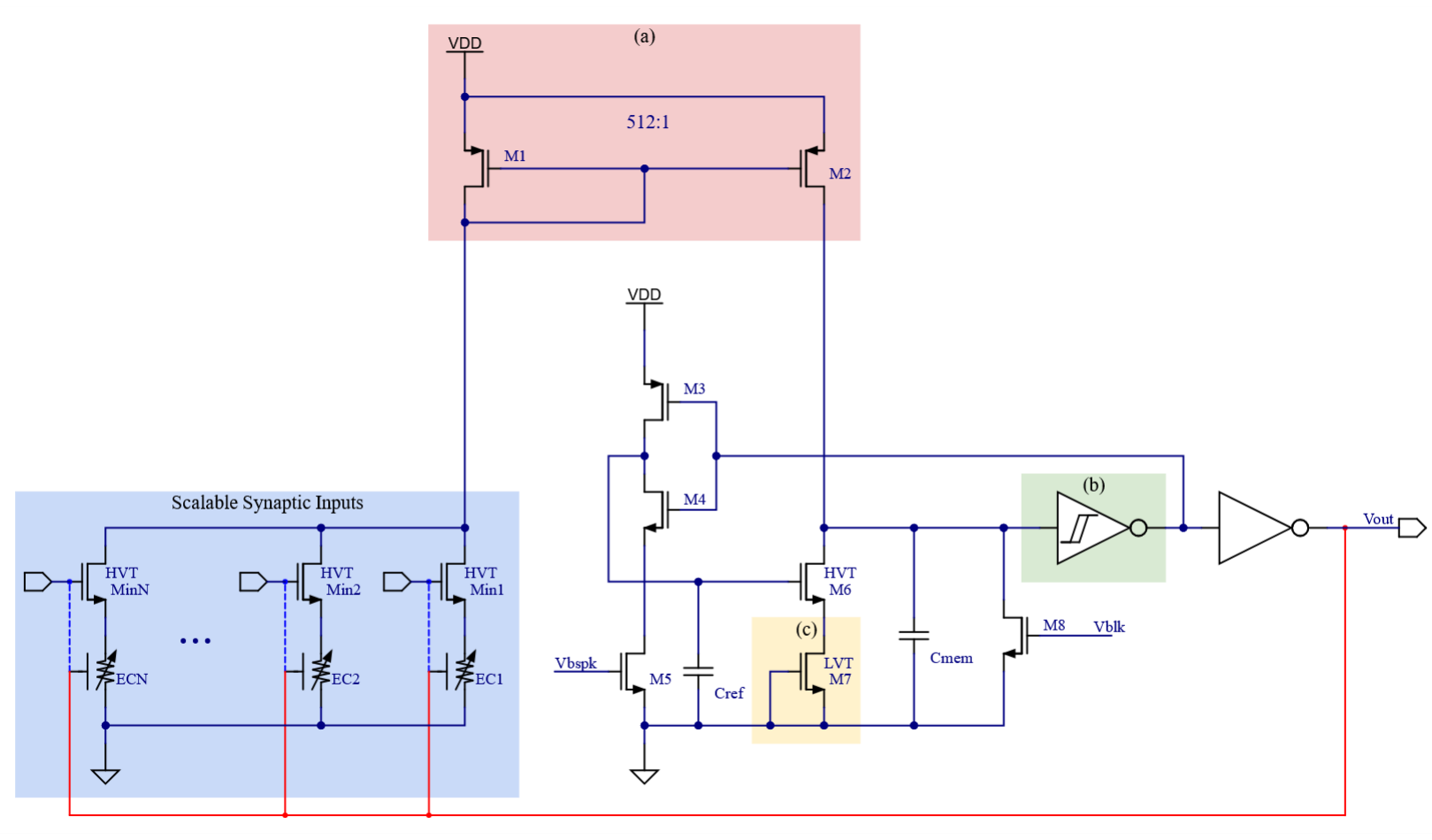}
    \caption{LIF neuron circuit based on the topology introduced in \cite{jooq_high-performance_2023}. Our modifications include: (a) a high current mirroring ratio at the synaptic input stage (b) replacement of a traditional CMOS inverter in the output stage with a CMOS schmitt trigger and (c) a high-resistance transistor placed in the membrane voltage reset path. Similarly to the current-starving topology, STP can be implemented by either connecting ECRAM gates to their corresponding synapses to emulate synaptic facilitation (shown by blue dashed lines) or connecting all ECRAM gates to the output to emulate intrinsic excitability modulation (shown by red solid lines).}
    \label{fig:prior_LIF}
    \end{mdframed}
\end{figure}

Our implementation of the capacitive refractory LIF design is shown in figure \ref{fig:prior_LIF}. In this architecture, synaptic current spikes are mirrored by M1 and M2 onto the membrane capacitor $C_{mem}$, gradually charging it and increasing the voltage $V_{mem}$. When $V_{mem}$ exceeds the turn-off threshold of the Schmitt trigger inverter in the output stage, the secondary inverter turns on and generates a high logic level at the circuit output. When this occurs, the skewed current-starved inverter circuit formed by M3-M5 turns on, charging the capacitor $C_{ref}$ and subsequently turning on M6. This allows $C_{mem}$ to discharge through M6 and M7, resetting $V_{mem}$ to ground. When $V_{mem}$ drops below the Schmitt trigger turn-on threshold, the output spike ends and the circuit is fully reset. 

The bias on the gate of M5, $V_{bspk}$ limits the current flow discharging $C_{ref}$, allowing M6 to discharge $C_{mem}$ for an additional time period after the spike to emulate the biological refractory period. Leakage functionality is added through M8, which is biased in the subthreshold region by an external voltage $V_{blk}$ to provide a constant low-conductance discharge path for $C_{mem}$.

The relatively large time constants associated with ECRAM volatile conductance dynamics pose an inherent challenge for direct integration with conventional CMOS neuron circuits. While CMOS circuitry typically operates in the nanosecond to microsecond regime, our fabricated ECRAM devices do not exhibit a significant increase in conductance when voltage pulses much shorter than a millisecond are applied to the gate. To accommodate these dynamics, we introduce three low-cost modifications to the original LIF neuron presented by Jooq et al. in order to intentionally reduce its operating speed to the millisecond timescale.

First, the input current mirror ratio is increased so that long-duration synaptic current pulses do not immediately trigger firing events without requiring proportional scaling of $C_{mem}$. The second modification replaces the standard inverter in the output stage with a Schmitt trigger inverter, ensuring that once $V_{mem}$ crosses the firing threshold, it must fall to the lower hysteresis threshold before the output spike terminates.  The third modification introduces a high-resistance FET in the reset path (M7), which significantly increases the reset path time constant and correspondingly extends the output spike duration. By connecting the gate and source of this transistor together, the device behaves as an ultra-high resistance defined primarily by leakage current.

An LVT device is chosen for M7 to deliberately increase leakage compared to the standard-threshold pull-up device M2, ensuring that the circuit can effectively return $V_{mem}$ to $VSS$ after a firing event. The total duration of the output pulse is determined by the reset-path time constant and the hysteresis window of the Schmitt trigger, allowing its width to be tuned through either parameter. In our implementation, we design the hysteresis of the Schmitt trigger to result in a 1 ms output spike, given the nominal resistance of M7.

These results demonstrate that volatile ECRAM dynamics can be leveraged to implement synaptic facilitation and intrinsic excitability modulation across multiple neuron circuit architectures, enabling STP functionality without requiring additional circuitry


\section{Device and Circuit-level Evaluation}\label{section:results}

This section evaluates the proposed device–circuit framework through validation of the ECRAM device model and demonstration of circuit-level STP mechanisms enabled by volatile conductance dynamics, with the resulting network-level implications analyzed in Section \ref{section:network}.

\subsection{Experimental Validation of ECRAM Device Model}
\begin{figure}
    \centering
   \begin{mdframed}[linewidth=0.35pt,linecolor=black!35]
   \centering
    \includegraphics[width=0.5\textwidth]{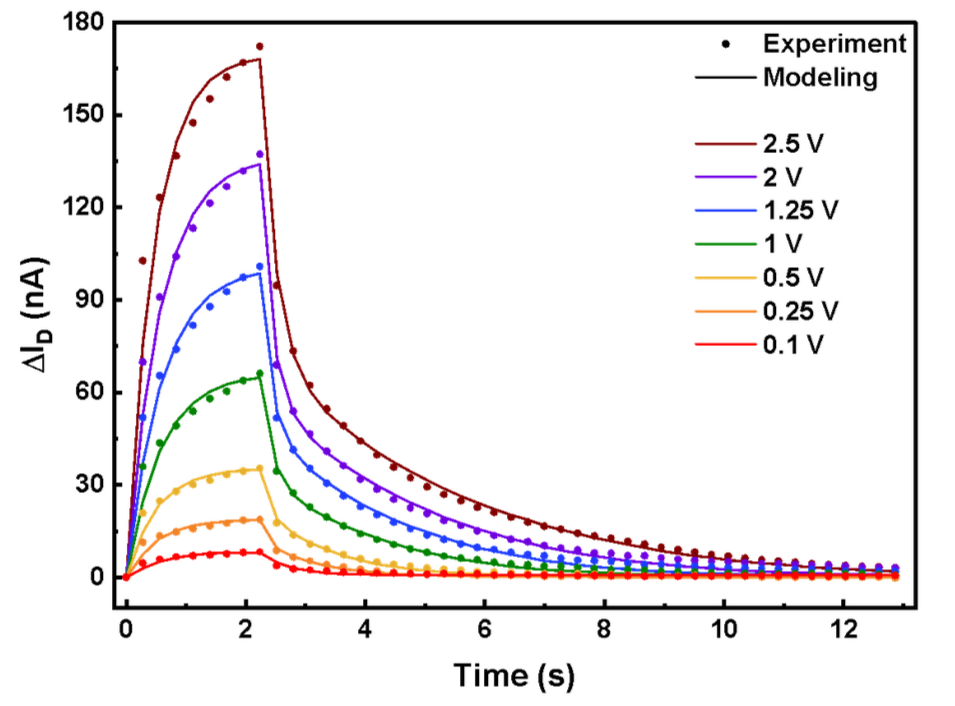}
    \caption{Measured volatile current dynamics of fabricated ECRAM devices in response to a 2.5s gate voltage pulse 2.5 applied at varying amplitudes. The model behavior (solid lines) closely matches ($R^2=0.97-0.99$) the experimental data (markers) for all input pulses, demonstrating an accurate recreation of ECRAM volatile dynamics in a simulated environment.}
    \label{fig:ecram_model_results}
    \end{mdframed}
\end{figure}

We begin by validating the physics-based compact model developed in this work by comparing simulated device behavior against experimental measurements. The volatile response of the device was characterized by measuring the transient drain current following the application and removal of a single gate voltage pulse. Figure \ref{fig:ecram_model_results} compares experimentally measured and simulated drain current transients for representative gate amplitudes. The model accurately reproduced both the magnitude and temporal evolution of the conductance increase during gate biasing, as well as the subsequent relaxation back to the baseline state after pulse removal. Strong quantitative agreement is observed across the measured voltage range ($R^2 = 0.97$--$0.99$), confirming that the extracted ionic relaxation parameters capture the underlying electric double-layer dynamics. 

Importantly, the model returns to the original conductance state after pulse removal, ensuring that volatile operation does not induce unintended nonvolatile drift. Nonvolatile operation is incorporated using experimentally demonstrated conductance states reported in our prior work \cite{han_energy_2025}. The model therefore preserves the programmed baseline conductance while allowing volatile modulation to be superimposed, enabling transient short-term facilitation effects without altering long-term state retention. This separation of volatile and nonvolatile mechanisms is essential for the circuit behaviors analyzed in the following sections.

\subsection{Circuit-Level Demonstration of ECRAM-enabled STP}
\begin{figure}
    \centering
    \begin{mdframed}[linewidth=0.35pt,linecolor=black!35]
    \centering
    \includegraphics[width=0.7\textwidth]{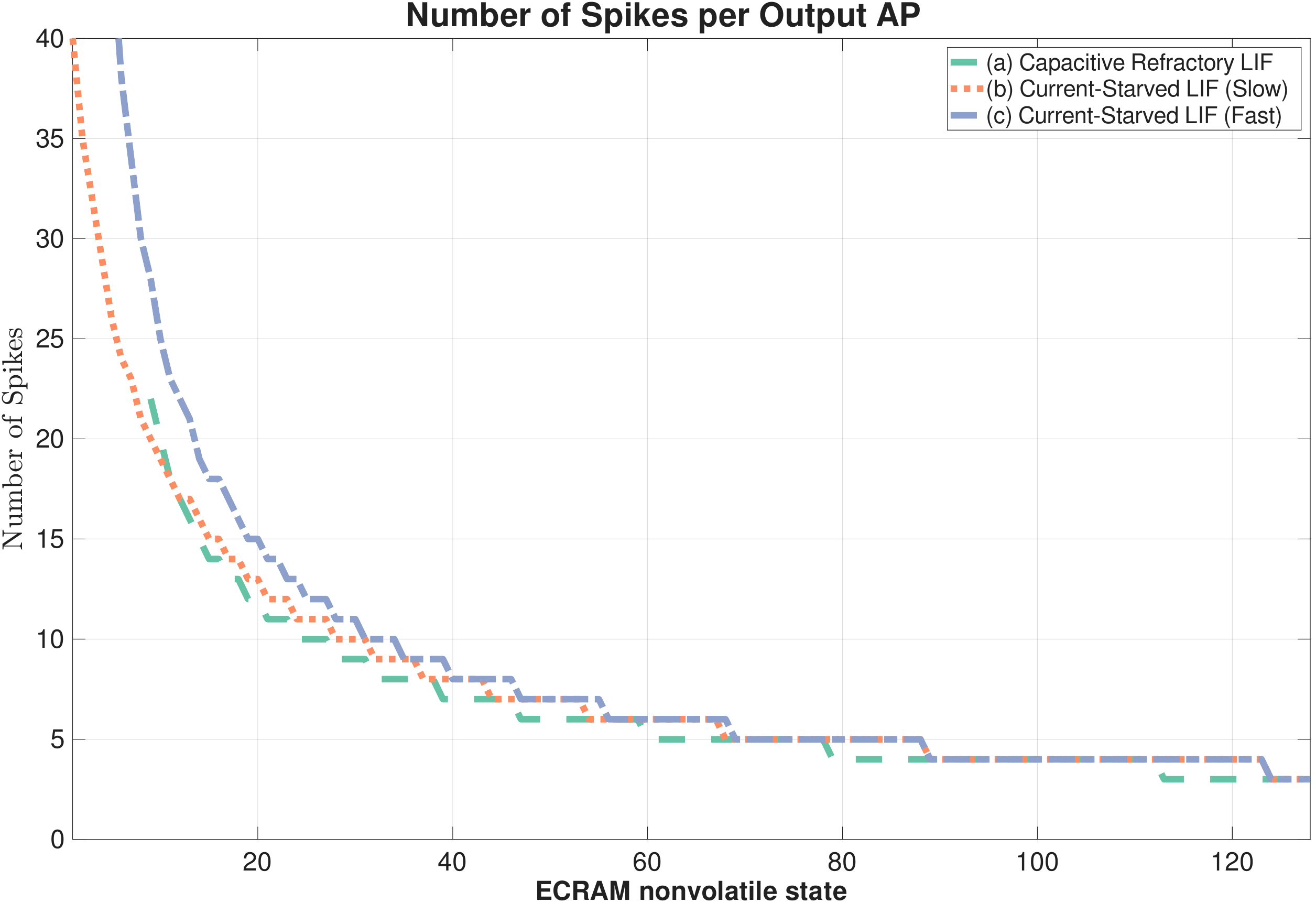}
    \caption{Number of input spikes required for the capacitive refractory topology (a) and delay-feedback topology at slow (b) and fast (c) timescales to generate an output spike at each nonvolatile ECRAM conductance state. As nonvolatile conductance of the ECRAM device decreases, the synaptic current generated by each input spike is reduced, resulting in a smaller change in membrane voltage. As such a greater number of spikes are required to reach the neuron firing threshold.}
    \label{fig:numspikes}
    \end{mdframed}
\end{figure}

To evaluate the functionality of the proposed architectures, all circuits were implemented in the GlobalFoundries 55 nm CMOS process using 1.2V devices in conjunction with the experimentally validated Verilog-A ECRAM model described previously. A train of 1 ms duration, 1.2 V amplitude input pulses was applied to the input of each LIF neuron circuit at a rate of 100 Hz to model sustained synaptic activity. For the fast-timescale variant of the delay-feedback LIF, a 100 kHz pulse train with 1 $\mu s$ pulses was applied instead. The number of input spikes required to generate an output was recorded for each nonvolatile ECRAM state to show functionality over the full range of conductance states (figure \ref{fig:numspikes}). Additional simulations were performed with the ECRAM gate connected either to the synaptic input or to the neuron output in order to emulate synaptic facilitation and intrinsic excitability modulation, respectively. 

Under baseline nonvolatile operation, all circuit implementations exhibited the expected monotonic relationship between synaptic conductance and neuronal firing behavior. Specifically, the number of input spikes required to generate an output action potential increased as the ECRAM conductance decreased. Greater sensitivity to conductance variation was observed in the low-conductance regime, where a single conductance state change could alter the firing threshold by several input spikes. At higher conductance states, the neuron response becomes more stable, with multiple adjacent states producing the same spike count threshold.

While this behavior may appear redundant for a single synapse driven by uniform spike trains, the small differences in membrane voltage modulation between adjacent states remain computationally meaningful when neurons are embedded in larger networks, particularly when temporal variability in input activity and STP mechanisms are introduced. The impact of ECRAM-enabled STP on neuron firing behavior is summarized in Table \ref{tab:LIF_Comparison} and analyzed in detail in the following subsections.

\begin{table}[ht]
\centering
\caption{Comparison of LIF Architectures.}
\label{tab:LIF_Comparison}
\scriptsize
\setlength{\tabcolsep}{1.8pt} 
\renewcommand{\arraystretch}{3}
\setlength{\heavyrulewidth}{0.05em}
\setlength{\lightrulewidth}{0.05em} 
\begin{tabularx}{\linewidth}{@{} >{\raggedright\arraybackslash}X ccccccccc @{}}
\toprule
\thead{Architecture\\ \\ \\} & 
\thead{ECRAM\\State\\ \\} & 
\thead{$\Delta$$V_{mem}$\\from\\1st Input} & 
\thead{Output\\Pulse\\Width} & 
\thead{Energy per\\Spike (No\\Facilitation)} & 
\thead{Energy per\\Spike (w/\\Facilitation)} & 
\thead{Spikes to\\generate\\AP} & 
\thead{Facilitation\\Type \\ \\} & 
\thead{Time to\\Partial\\Facilitation} & 
\thead{Time to\\Full\\Facilitation} \\ \midrule

\multirow{2}{=}{Delay-Feedback - Long Timescale (50 Hz)} & \multirow{2}{*}{60} & \multirow{2}{*}{0.129 V} & \multirow{2}{*}{1.216 ms} & \multirow{2}{*}{2.438 nJ} & \multirow{2}{*}{2.439 nJ} & \multirow{2}{*}{7} & Neuronal & -- & -- \\ \cmidrule(lr){8-10} & &  &  &  &  &  & Synaptic & -- & 540 ms \\ \midrule

\multirow{2}{=}{Delay-Feedback - Long Timescale (100 Hz)} & \multirow{2}{*}{60} & \multirow{2}{*}{0.129 V} & \multirow{2}{*}{1.242 ms} & \multirow{2}{*}{2.396 nJ} & \multirow{2}{*}{2.422 nJ} & \multirow{2}{*}{6} & Neuronal & 1.25 s & -- \\ \cmidrule(lr){8-10} & &  &  &  &  &  & Synaptic & -- & 170 ms \\ \midrule

\multirow{2}{=}{Delay-Feedback - Long Timescale (200 Hz)} & \multirow{2}{*}{60} & \multirow{2}{*}{0.129 V} & \multirow{2}{*}{1.237 ms} & \multirow{2}{*}{2.603 nJ} & \multirow{2}{*}{2.604 nJ} & \multirow{2}{*}{5} & Neuronal & 170 ms & 280 ms \\ \cmidrule(lr){8-10} & &  &  &  &  &  & Synaptic & -- & 70 ms \\ \midrule

\multirow{2}{=}{Delay-Feedback - Short Timescale (100 kHz)} & \multirow{2}{*}{82} & \multirow{2}{*}{0.169 V} & \multirow{2}{*}{1.136 $\mu$s} & \multirow{2}{*}{1.954 pJ} & \multirow{2}{*}{1.962 pJ} & \multirow{2}{*}{5} & Neuronal & 2.89 ms & 3.29 ms \\ \cmidrule(lr){8-10} & &  &  &  &  &  & Synaptic & 2.19 ms & 3.72 ms \\ \midrule

\multirow{2}{=}{Capacitive Refractory (100 Hz)} & \multirow{2}{*}{112} & \multirow{2}{*}{0.289 V} & \multirow{2}{*}{1.195 ms} & \multirow{2}{*}{8.213 nJ} & \multirow{2}{*}{8.222 nJ} & \multirow{2}{*}{4} & Neuronal & 300 ms & 2.73 s \\ \cmidrule(lr){8-10} & &  &  &  &  &  & Synaptic & 100 ms & 240 ms \\ \bottomrule
\end{tabularx}
\end{table}

\subsubsection{Delay-Feedback LIF Neuron - Long Timescale}
We repeated the transient simulation of our novel LIF circuit with the ECRAM gate connected to the LIF output for three different frequencies of input spiking: 50 Hz, 100 Hz, and 200 Hz. Each output spike biases every synapse towards higher current flow, effectively lowering the neuron firing threshold and implementing intrinsic excitability modulation. In each case, the circuit required approximately 2.4-2.6 nJ of energy to produce a 1.2 ms output spike, representing only a negligible increase in energy consumption compared to operation without STP. Each output spike provided a voltage pulse to the gate of every synaptic ECRAM device, reducing the synaptic resistance by approximately 6.8 $k\Omega$ in nonvolatile conductance state 60. 

As the frequency of input activity increased from 50 Hz to 200 Hz, the number of spikes required to initiate a firing event decreased from 7 to 5, demonstrating the dependence of output behavior not only on the number of input spikes but also the temporal dynamics of their arrival at the synapse. Input frequency also had an impact on the degree of facilitation observed. At 50 Hz, no amount of output activity was able to reduce the number of spikes needed to fire the neuron. At 100 Hz, only partial facilitation was achieved. We define partial facilitation as a firing pattern that emerges when the ECRAM conductance lies near the threshold for a reduced spike count, resulting in alternating firing events requiring 4 and 5 input spikes. This can be attributed to the fact that the ECRAM conductance partially recovers between output spikes due to volatile ion dynamics, causing it to oscillate around the threshold for increased firing frequency. 

\begin{figure}
    \centering
    \begin{mdframed}[linewidth=0.35pt,linecolor=black!35]
    \centering
    \includegraphics[width=\textwidth]{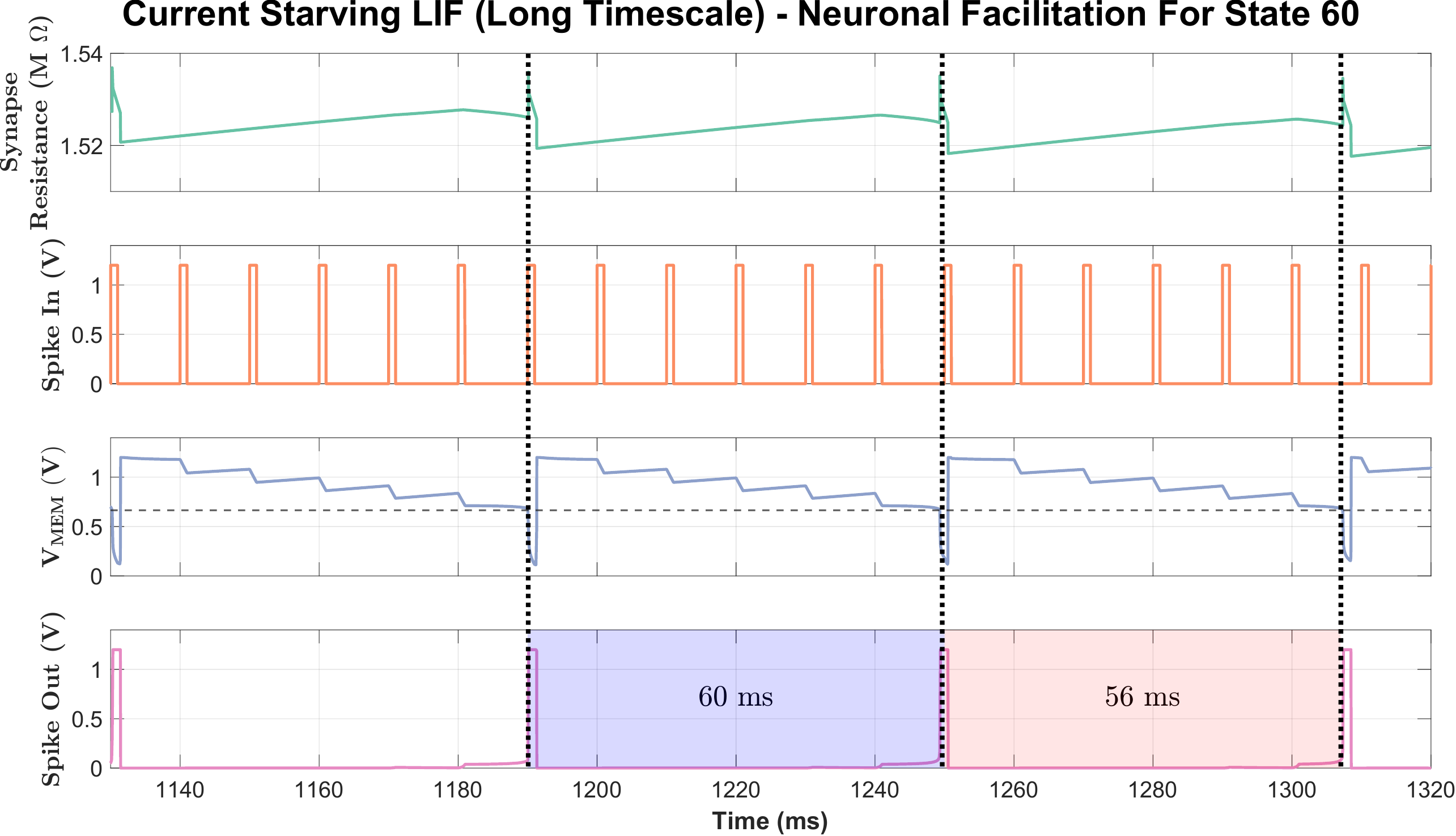}
    \caption{Increased activity of the delay-feedback LIF neuron circuit with a synapse in state 60. After 1.2 seconds of repeated input APs, every firing event takes one fewer input to trigger than the initial state, showcasing neuronal facilitation. Highlighted in blue is the output triggering after 6 input stimuli, and highlighted in red is the output triggering after 5 input stimuli.}
    \label{fig:design_2_slow_neuronal_facilitation}
    \end{mdframed}
\end{figure}

The first facilitated output spike was observed after 1.2 seconds of input activity (figure \ref{fig:design_2_slow_neuronal_facilitation}), demonstrating an increase in excitability driven by repeated spiking activity. For the 200 Hz case, facilitation emerges significantly faster, with the first facilitated spikes appearing at 170 ms, while full facilitation (defined as every output spike requiring one fewer input spike than the initial condition)  occurs after only 280 ms of sustained input spikes. These results demonstrate a circuit-level implementation of intrinsic excitability modulation and highlight how proposed neuron can detect temporal dynamics in synaptic activity, enabling richer computational behavior without additional circuit components.

\begin{figure}
    \centering
    \begin{mdframed}[linewidth=0.35pt,linecolor=black!35]
    \centering
    \includegraphics[width=\textwidth]{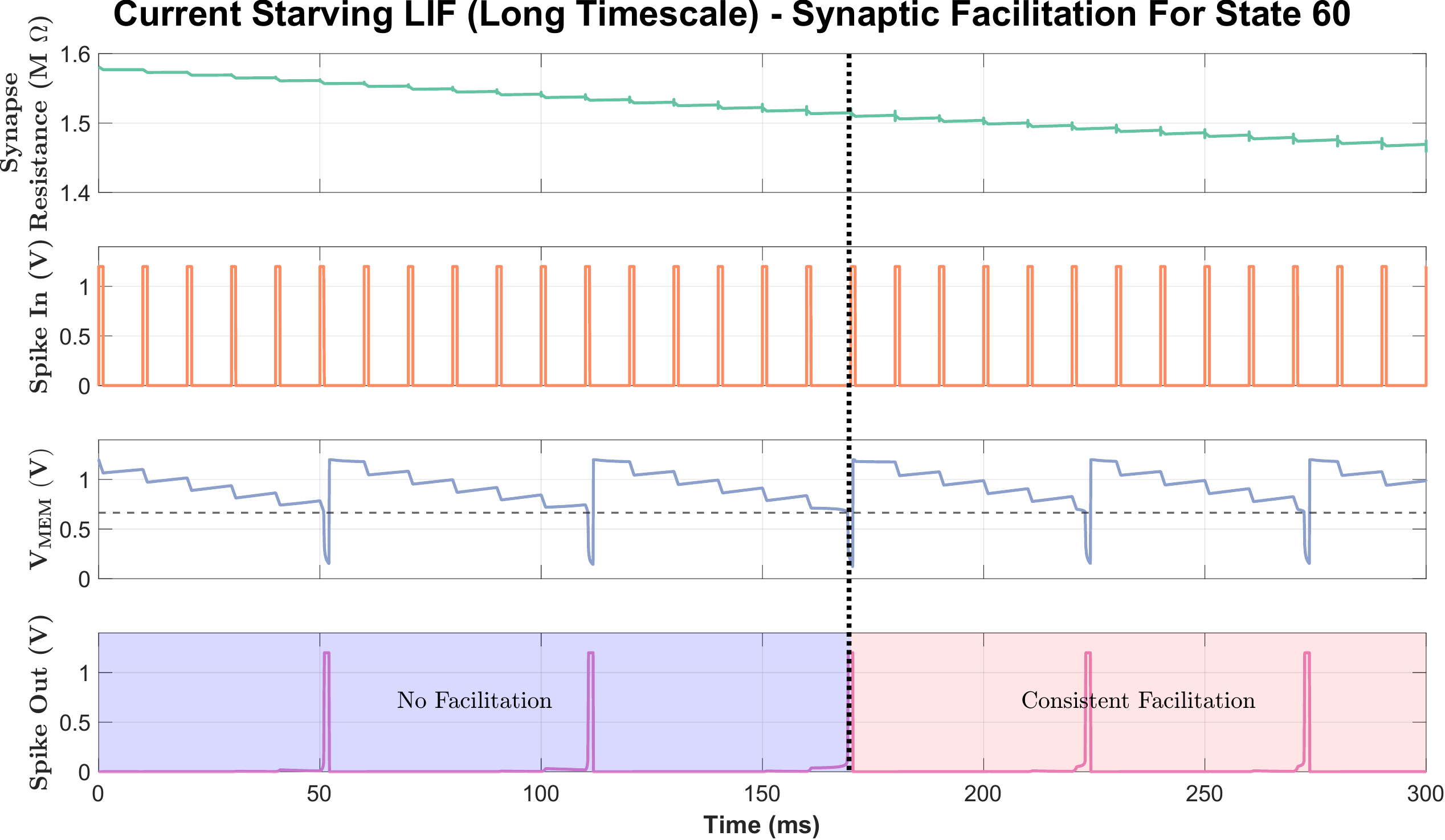}
    \caption{Increased activity of the current-starving LIF circuit with a synapse in state 60. After 170 milliseconds of repeated input APs, every firing event takes one fewer input to trigger than the initial state, showcasing synaptic facilitation. Highlighted in blue is the output triggering after 6 input stimuli, and highlighted in red is the output triggering after 5 input stimuli.}
    \label{fig:design_2_slow_synaptic_facilitation}
    \end{mdframed}
\end{figure}

Next, synaptic facilitation was demonstrated by connecting the spiking input of the synapse to its corresponding ECRAM gate and repeating the input pulse train simulation at 50, 100, and 200 Hz. In this configuration, the ECRAM gate is pulsed at a far higher frequency than the previous configuration, and as such facilitation effects emerge after a comparatively short amount of time. The higher frequency of gate pulsing also eliminates the intermittent facilitation regime observed previously, as the ECRAM gate conductance is driven more strongly past the threshold required for facilitated output behavior. Figure \ref{fig:design_2_slow_synaptic_facilitation} demonstrates that for the 100 Hz case, every firing event after 170 ms of input activity requires 5, rather than the initial 6, input spikes to occur. As in the intrinsic excitability case, increasing the frequency of input activity reduces the time required for facilitation to emerge, with the 50 Hz case requiring 540 ms and the 200 Hz case requiring only 70 ms of input spiking to exhibit facilitated behavior. 

Since each ECRAM gate can be individually connected to its corresponding synaptic input, this configuration demonstrates true synaptic-level STP, where only synapses experiencing high activity will undergo facilitation. In contrast to the neuron-wide excitability modulation described previously,  facilitation is locally tunable at the synaptic level, enabling richer spatial and temporal dynamics within the LIF scale - as demonstrated in Section \ref{section:network}.

\subsubsection{Delay-Feedback LIF - Short Timescale}
When modified to operate at the microsecond scale, intrinsic excitability modulation and synaptic facilitation behaviors remain present when a pulse train is applied to the circuit input. For each 1 $\mu s$ gate voltage pulse, the ECRAM device in the LIF synapse decreased its resistance by approximately 1.5 k$\Omega$, enabling STP as a function of repeated pulses. Initially, this configuration required 5 input spikes to consistently produce a firing event, but as synaptic conductance increases with repeated output activity, intermittent firing events occur after only 4 input spikes. Each of these intermittent facilitated outputs serves to drive the ECRAM conductance higher, eventually resulting in the synapse consistently requiring 4 input spikes to initiate a firing event after 3.3 ms of consistent input activity.

Synaptic facilitation was implemented in the same manner as in the slower timescale simulation. The same facilitation behavior was observed, but within a shorter time window due to the increased frequency of ECRAM gate pulsing. Only 2.18 ms of input activity is required before the circuit begins producing output spikes every 4 inputs, as opposed to the initial  5-input firing threshold. An important result of operating at a faster timescale is the significantly reduced energy required to produce a firing event, with approximately only 2 pJ required to produce an output AP both with and without STP implemented. This reduction arises from the substantially shorter duration of the feedback path activity during each firing event. Additionally, the ability to scale down the circuit to the microsecond timescale both reduces the area requirements for components such as capacitors and current mirroring transistors and vastly improves computational speed and throughput. Consequently, further implementations may benefit from ECRAM devices specifically designed to minimize time constants associated with volatile ion dynamics.

\subsubsection{Variability Analysis}
\input{variability.tex}

\begin{figure}
    \centering
    \begin{mdframed}[linewidth=0.35pt,linecolor=black!35]
    \centering
    \includegraphics[width=\textwidth]{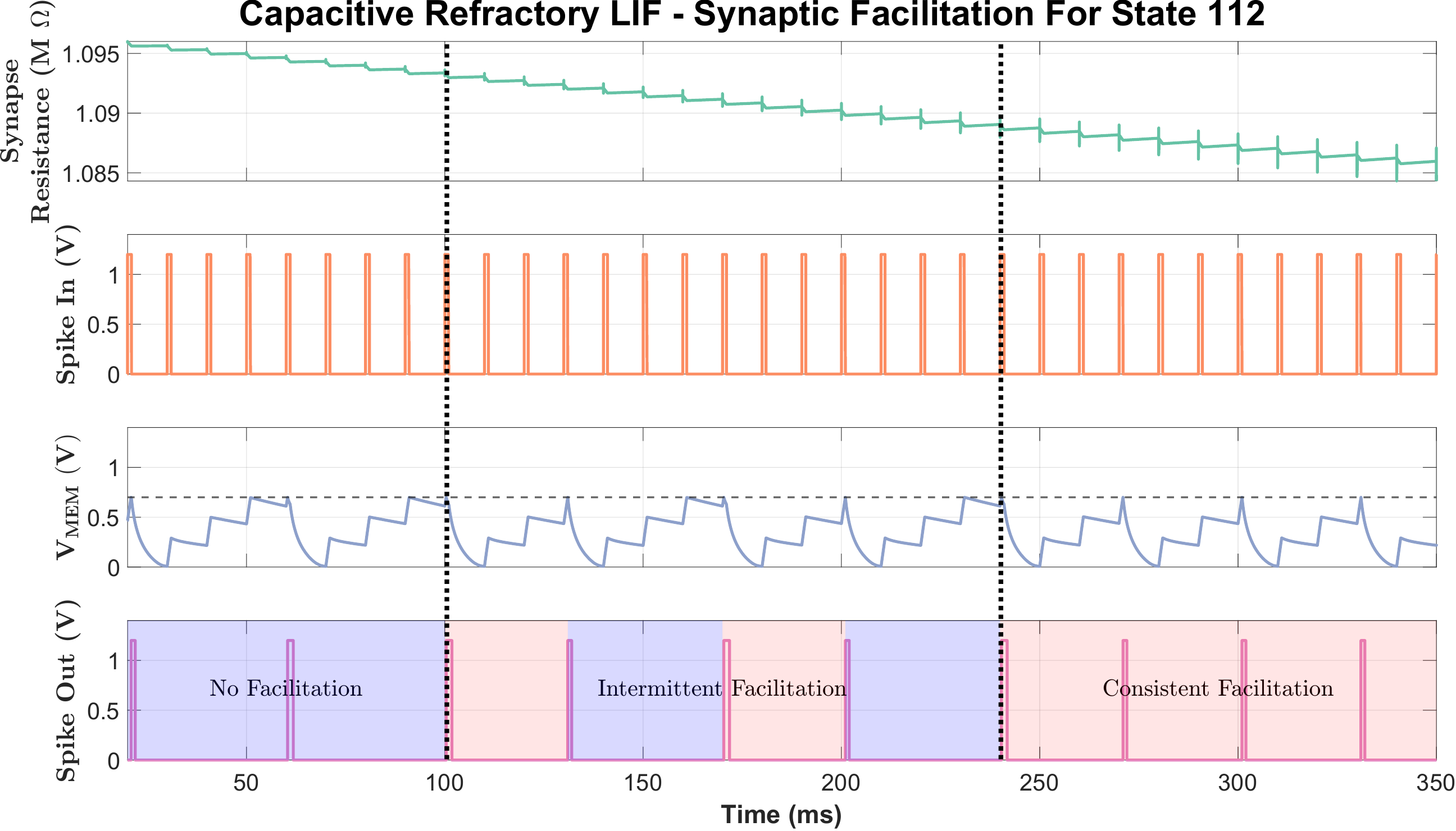}
    \caption{Synaptic facilitation of the capacitive refractory topology with a synapse in conductance state 112. During the first 100 ms, no facilitation occurs and each output spike requires 4 input spikes. Between 100 and 240 ms, a firing pattern of 3-4-3-4 emerges as the synaptic resistance passes through a near-threshold state. Finally, after 240 ms of repeated input activity, the resistance decreases sufficiently to allow each output spike to occur after only 3 input spikes.}
    \label{fig:design_1_synaptic_facilitation}
    \end{mdframed}
\end{figure}

\begin{figure}
    \centering
    \begin{mdframed}[linewidth=0.35pt,linecolor=black!35]
    \centering
    \includegraphics[width=\textwidth]{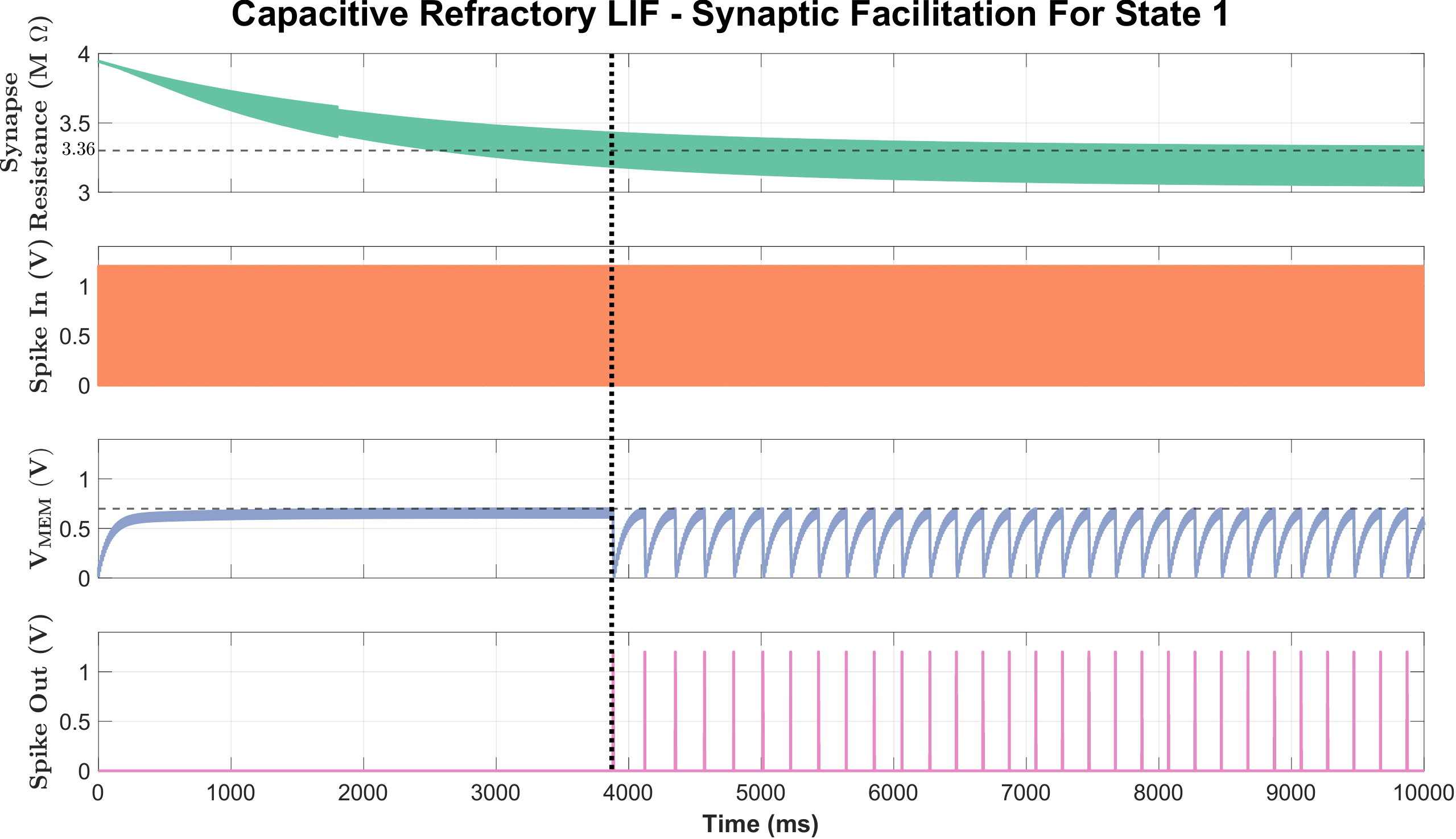}
    \caption{Activation of a low-conductance synapse in ECRAM conductance state 1 after repeated facilitation from input APs. Initially, membrane leakage dominates synaptic current such that $V_{mem}$ is unable to reach the firing threshold. As repeated input spikes induce synaptic facilitation, the synaptic conductance gradually increases, allowing the current to eventually overcome leakage and begin generating output APs.}
    \label{fig:design_1_neuron_activation}
    \end{mdframed}
\end{figure}

\subsubsection{Capacitive Refractory LIF}

When initial simulations were performed on the capacitive refractory LIF architecture, the observed behavior was largely the same as in the previous two topologies. The number of spikes required to generate an output ranges from 3 to 22 between ECRAM states 128 and 9, respectively. However, an interesting case appears for states 1-8 in which the synaptic conductance is so low that leakage currents dominate and no number of spikes at the specified 100 Hz rate can trigger a firing event. This is inherently problematic, as in a network scenario a low-conductance synapse can be used to bias a higher-conductance synapse towards firing sooner, or two low-conductance synapses can work together to generate a single output. Further experiments also demonstrated that the introduction of STP can exploit these states to create a behavior in which a baseline activity level must be achieved before a synapse "activates" and is able to trigger output APs independently.

Overall, the capacitive refractory topology requires slightly more energy to produce an AP than the delay-feedback LIF, with approximately 8.2 nJ required per firing event. However, as with the previous topologies, the introduction of STP produced a negligible increase in energy per spike, since minimal dynamic current is required to switch the ECRAM device gate terminals embedded in the synapses.


When neuronal facilitation is introduced, several states begin exhibiting modified behavior after a number of output spikes. Initially, with the synaptic ECRAM in state 112, four spikes are required to generate an output. It was observed that the resistance of the ECRAM device decreases by approximately 600 $\Omega$ each time an output AP is generated, with volatile ion dynamics causing the conductance to recover slightly between each spike. After eight APs within 300 ms, the synapse becomes sufficiently facilitated that an output can be generated with only three input spikes. At this point, a pattern of intermittent facilitated output spikes occurs until approximately 2.75 seconds of repetition, at which point the synaptic conductance rises enough to trigger an AP consistently every three input events. This result shows the implementation of STP in the form of intrinsic excitability modulation in another LIF architecture, with the entire neuron increasing in input sensitivity as repeated APs are generated. 

Throughout this experiment, no change in the nonvolatile ECRAM state is induced, meaning that given a sufficiently long duration without synaptic activity, the circuit relaxes to its initial state. Because facilitation in this configuration depends on output spikes, ECRAM states 1-8 remain unable to trigger firing, as no output APS are generated to drive facilitation. However, the following experiment with synaptic facilitation demonstrates how such low conductance synapses can be activated under appropriate conditions.

Synaptic facilitation was implemented by connecting incoming synaptic spike signals to both the standard synaptic input and the corresponding ECRAM gate terminal. As a result, each incoming spike temporarily modulates the ECRAM conductance, implicitly biasing the neuron towards firing through activity at that synapse – thus enabling synaptic facilitation. In this configuration, high conductance synaptic states such as 112 continue to show increased activity with repeated spikes, but the facilitation occurs much faster due to the increased frequency of ECRAM gate stimulation. The first reduced-input AP occurs after 100 ms, and consistent reduced-input APs begin after 240 ms of input activity. This configuration provided a clear visualization of the intermittent facilitation behavior seen across all topologies (figure \ref{fig:design_1_synaptic_facilitation}). 

With facilitation occurring based on input activity rather than output activity, this configuration can also activate previously idle low-conductance synapses. For example, under the given input conditions a synapse in state 1 remains fully inactive for nearly 4 seconds of input activity, but is subsequently biased by synaptic facilitation to begin generating APs consistently every 24 input spikes (figure \ref{fig:design_1_neuron_activation}). This behavior presents an interesting opportunity for advanced neuron-level computation with minimal additional area overhead and extremely low power required for ECRAM gate switching. 

\textit{Overall, these results demonstrate that the proposed STP methodologies are not limited to a single neuron circuit topology or operating timescale, and can be applied to a broad class of ECRAM-based neuromorphic architectures. }

\section{Network-Level and Computational Analysis}
\label{section:network}

The circuit-level results presented above demonstrate that volatile ECRAM dynamics enable STP mechanisms such as synaptic facilitation and intrinsic excitability modulation. To explore how these device- and circuit-level behaviors influence computation at the network scale, we implemented an abstracted model of the proposed synaptic dynamics within a spiking neural network framework. For network-level analysis, we implemented a synaptic facilitation/depression model along with LIF neurons in snnTorch. This abstraction captures the essential temporal dynamics observed in the hardware.

STP plays a key role in many neural computations, including temporal pattern detection, coincidence detection, adaptive filtering of spike trains, and activity-dependent recruitment of synapses. The following model allows us to study how ECRAM-enabled facilitation and depression influence these behaviors within a network of LIF neurons.

The LIF neuron with facilitating synapses is modeled as
\begin{equation}
    \tau_{s}\frac{\mathrm{d}s}{\mathrm{d}t}=-s+\tau_{s}\sum\limits_{i}w^{\prime}_{i}(t)\sum\limits_{k}\delta(t_{i,k}-t)
\end{equation}
\begin{equation}
    \tau_{f}\frac{\mathrm{d}w_{i}^{\prime}}{\mathrm{d}t}=-(w_{i}^{\prime}-w_{i})+\tau_{f}\Delta w_{i}^{\prime}\sum\limits_{k}\delta(t_{i,k}-t)
\end{equation}
where $s$ represents the (unitless) membrane variable with time constant $\tau_{s}$, $w$ and $w^{\prime}$ are the base and effective synaptic efficacies, $t_{i,k}$ represents the spike train for neuron $i$, and $\delta(\cdot)$ is the Dirac delta function.  When the membrane potential reaches a threshold $\theta$, it resets to zero, and the neuron emits a spike. Note that we could easily add a second ECRAM device to each synapse to emulate activity-dependent short term depression (STD).  The combined effect of facilitation and STD is modeled as:
\begin{equation}
\tau_{f}\frac{\mathrm{d}f_{i}}{\mathrm{d}t} = -f_{i} + \tau_{f}\Delta f_{i}\sum_{k}\delta(t_{i,k}-t)
\label{eqn:facil}
\end{equation}
\begin{equation}
\tau_{d}\frac{\mathrm{d}d_{i}}{\mathrm{d}t} = -d_{i} + \tau_{d}\Delta d_{i}\sum_{k}\delta(t_{i,k}-t)
\label{eqn:depress}
\end{equation}
\begin{equation}
w'_{i}(t) = 
\begin{cases}
\max(w_{i}+f_{i}(t)-d_{i}(t),0) & \text{if } w_{i}>0 \\
\min(w_{i}+d_{i}(t)-f_{i}(t),0) & \text{if } w_{i}<0
\end{cases}
\label{eqn:effweight}
\end{equation}

\begin{figure}[!ht]
\vspace{2mm}
\begin{mdframed}[linewidth=0.35pt,linecolor=black!35]
\centering
\subfigure[]{
\includegraphics[width=0.7\textwidth]{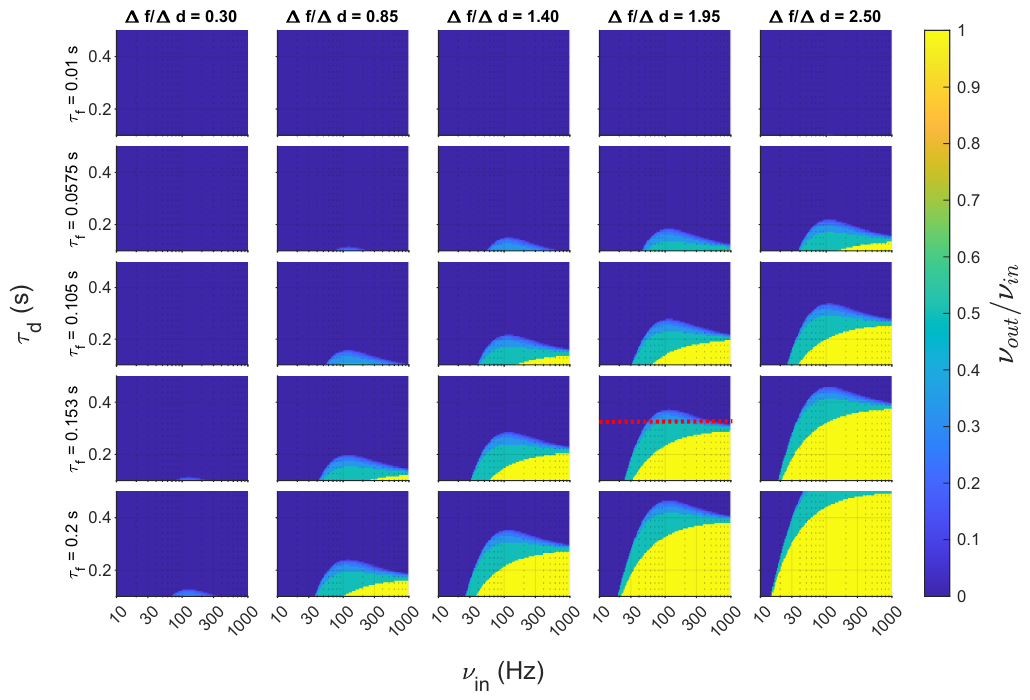}}
\subfigure[]{
\includegraphics[width=0.7\textwidth]{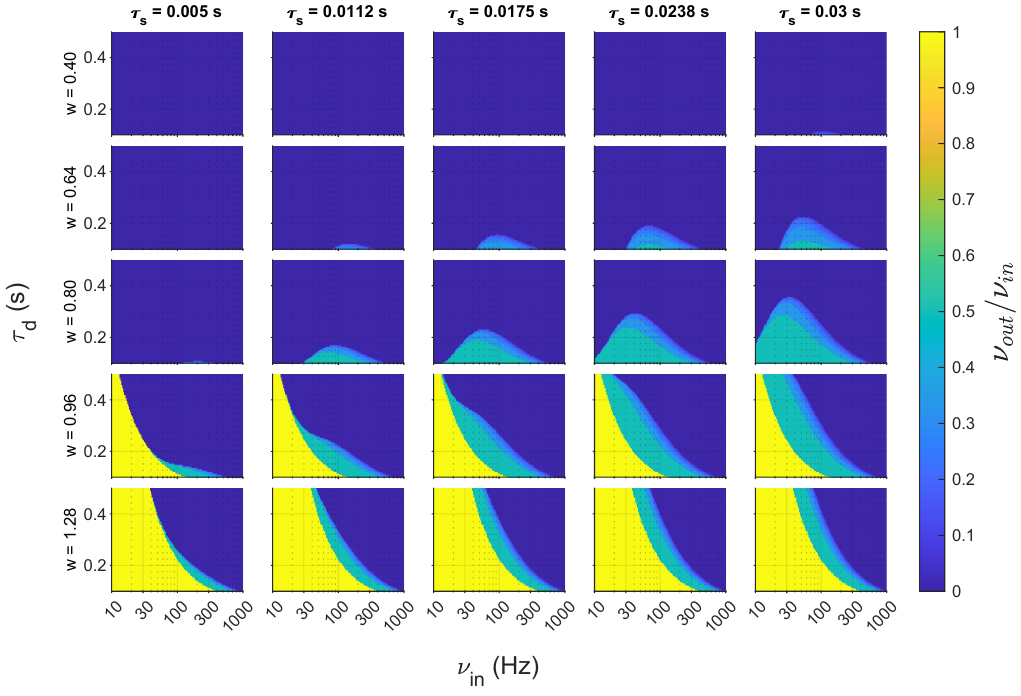}}

\caption{Frequency response of an LIF neuron (Phase space analysis) with ECRAM-based facilitating/depressing synapse.  Each panel shows the frequency response $\nu_{out}/nu_{in}$ vs. $\tau_{d}$ (vertical axis) and $\nu_{in}$ (horizontal axis) illustrating how STP dynamics enable frequency-selective temporal filtering of spike trains.  Rows and columns in (a) correspond to different values of $\tau_{f}$ and $\Delta f/\Delta d$, respectively, with $w=0.6$, $\tau_{s}=0.015$ s, and $\Delta d=0.03$.  Rows and columns in (b) correspond to different values of $w$ and $\tau_{s}$, respectively, with $\tau_{f}=0.05$ s, $\Delta d=0.03$, and $\Delta f/\Delta d=1.2$.  In all cases, $\theta = 0.8$.}
\label{fig:freqresp}
\end{mdframed}
\end{figure}

\begin{figure}
\begin{mdframed}[linewidth=0.35pt,linecolor=black!35]
    \centering
    \includegraphics[width=0.75\linewidth]{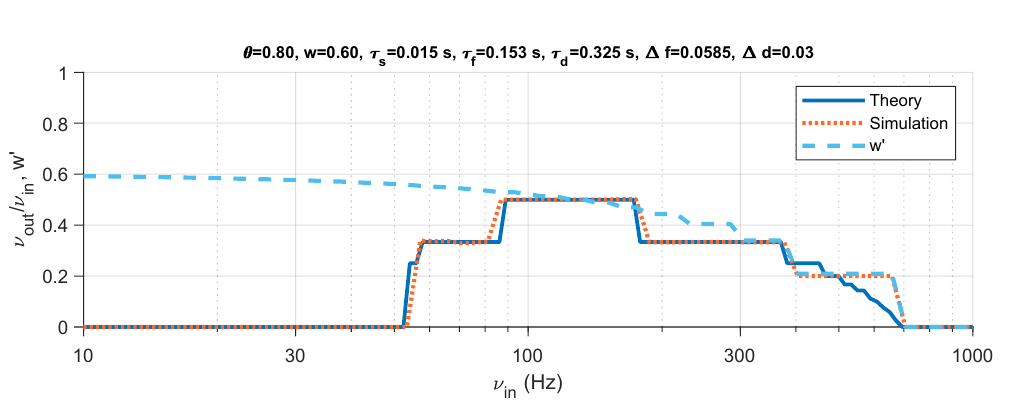}
    \caption{Comparison of analytical and simulated frequency responses for a 1-d slice through parameter space of the ECRAM-based synapse with facilitation and depression capabilities. In this parameter regime, the combined effects of synaptic dynamics and membrane leakage produce a band-pass–like frequency response, where spiking occurs only within a limited range of input frequencies}
\label{fig:paramslice}
\end{mdframed}
\end{figure}

These equations reproduce the activity-dependent modulation of synaptic efficacy observed in the circuit simulations and enable analysis of how ECRAM-based STP influences computation in spiking neural networks. The inclusion of the dynamics discussed above in neuromorphic synapses enables relatively complex spatiotemporal processing in single synapses.  Without dynamic synapses, the same processing would require multi-synapse/multi-neuron networks.  For example, tuning of facilitation/depression parameters can be used to configure individual synapses as frequency filters.  The leaky membrane of the LIF acts as a high-pass filter, and the composition of the facilitating/depressing synapses with the LIF enables high pass, low pass, and band pass filtering.  Consider an LIF neuron a single input spiking at constant frequency $\nu_{in}$ or period $T_{in}=1/\nu_{in}$.  The facilitation and depression factors in (\ref{eqn:facil})-(\ref{eqn:depress}) can be written as discrete-time recurrences:
\begin{equation}
f_{k+1}=f_{k}\mathrm{e}^{-T/\tau_{f}}+\Delta f, \:\:\:\: d_{k+1}=d_{k}\mathrm{e}^{-T/\tau_{d}}+\Delta d
\end{equation}
Steady state is reached when $f_{k+1}=f_{k}$ (or $d_{k+1}=d_{k}$), leading to 
\begin{equation}
f(T)=\frac{\Delta f}{1-\mathrm{e}^{-T/\tau_{f}}}, \:\:\:\:, d(T)=\frac{\Delta d}{1-\mathrm{e}^{-T/\tau_{d}}}
\end{equation}
Additionally, the steady-state effective weight $w'(\nu_{in})$ at the input frequency $\nu_{in}$ can be found by using $f(T)$ and $d(T)$ in (\ref{eqn:effweight}).  The recurrence relation for the LIF can also be written as
\begin{equation}
s_{k+1}=s_{k}\mathrm{e}^{-T/\tau_{s}}+w'(\nu_{in}),
\end{equation}
and after $m$ pre-synaptic input spikes (when initially reset to 0), the membrane potential will be
\begin{equation}
s_{m}=w'(\nu_{in})\sum\limits_{k=0}^{m-1}\alpha^{k}=w'(\nu_{in})\frac{1-\alpha^{m}}{1-{\alpha}}
\end{equation}
where $\alpha\equiv\mathrm{e}^{-T/\tau_{s}}$.  The minimum number of spikes required for $s_{m}\ge\theta$ can then be solved as
\begin{equation}
m^{*}(\nu_{in})=\left\lceil\frac{\mathrm{ln}\left(1-\theta\frac{1-\alpha(\nu_{in})}{w'(\nu_{in})}\right)}{\mathrm{ln}(\alpha(\nu_{in}))}\right\rceil
\end{equation}
Finally, we can express the frequency response as
\begin{equation}
\frac{\nu_{out}}{\nu_{in}}=\frac{1}{m^{*}}.
\end{equation}
In figure \ref{fig:freqresp}, we show the frequency response described above for different parameters values for the time constants, facilitation/depression amounts, and synaptic weights.  From a single ECRAM-based synapse with appropriately-tuned dynamics, we can achieve multiple types of frequency filtering.  Moreover, synaptic dynamics allow each input to an LIF neuron to implement its own frequency-selective filtering, enabling richer temporal processing than relying on neuron dynamics alone.  A 1-d slice in the parameter space corresponding to the red dotted line in figure \ref{fig:freqresp}(a) was simulated in snnTorch to compare the theoretical and simulated frequency responses.  The results, shown in figure \ref{fig:paramslice}, confirm the accuracy of the theoretical model, with small errors stemming from the finite simulation time step (1 ms in this case).  Also illustrated in the plot is the effective weight $w'$ at each input frequency.  In this example, the high-pass filter properties of the soma enable spiking at mid frequencies, while the impact of synaptic depression dominates at high frequencies, creating a bandpass-like response.

These results demonstrate that volatile ECRAM dynamics enable synapses to function as tunable temporal filters, allowing compact neuromorphic circuits to perform frequency-selective and temporally structured computations directly within individual synapse–neuron pairs.

\section{Conclusion}

This work demonstrates that volatile non-equilibrium dynamics in ionically gated memristive devices can serve as a native computational primitive for neuromorphic hardware. Rather than treating transient conductance modulation in ECRAM devices as an undesirable artifact, we show that these dynamics can be directly exploited to realize biologically inspired STP through cross-layer co-design.

We introduce delay-feedback leaky integrate-and-fire (LIF) neuron architectures in which activity-dependent modulation of ECRAM synaptic conductance enables synaptic facilitation and intrinsic excitability modulation with negligible additional circuit overhead. Experimentally fabricated and characterized ECRAM devices were used to develop a compact behavioral model that captures the underlying transient conductance dynamics and enables accurate circuit-level simulation of the proposed mechanisms. Circuit simulations indicate that these dynamics can be realized with energy consumption on the order of \~2\,pJ per spike. Because volatile and nonvolatile behaviors coexist within the same device, the proposed approach allows short-term temporal dynamics and long-term synaptic weight storage to be implemented simultaneously within a unified synaptic element.

Circuit and network-level analysis further demonstrate that the interaction between ECRAM device dynamics and neuron integration enables frequency-selective spike processing, allowing individual synapses to act as tunable temporal filters. Importantly, the proposed mechanisms are not restricted to a single neuron implementation and are demonstrated across multiple LIF circuit topologies, suggesting broad compatibility with existing neuromorphic circuit architectures.

Overall, this work establishes volatile ECRAM dynamics as a practical substrate for embedding temporal computation directly within compact neuromorphic hardware. More broadly, it illustrates how embracing non-equilibrium device physics through device–circuit co-design can enable a new class of computational primitives, enabling compact neuromorphic circuits to perform temporal spike filtering directly within synapse–neuron pairs.

\ack{The authors would like to thank GlobalFoundries for their support through the University Partnership Program.}

\bibliographystyle{IEEEtran}
\footnotesize
\bibliography{IEEEabrv,myjournal}

\end{document}

%% file: variability.tex
\begin{figure}

    \begin{mdframed}[linewidth=0.35pt,linecolor=black!35]

    \centering
    \begin{minipage}{0.6\textwidth}
        \centering
        \includegraphics[width=\textwidth]{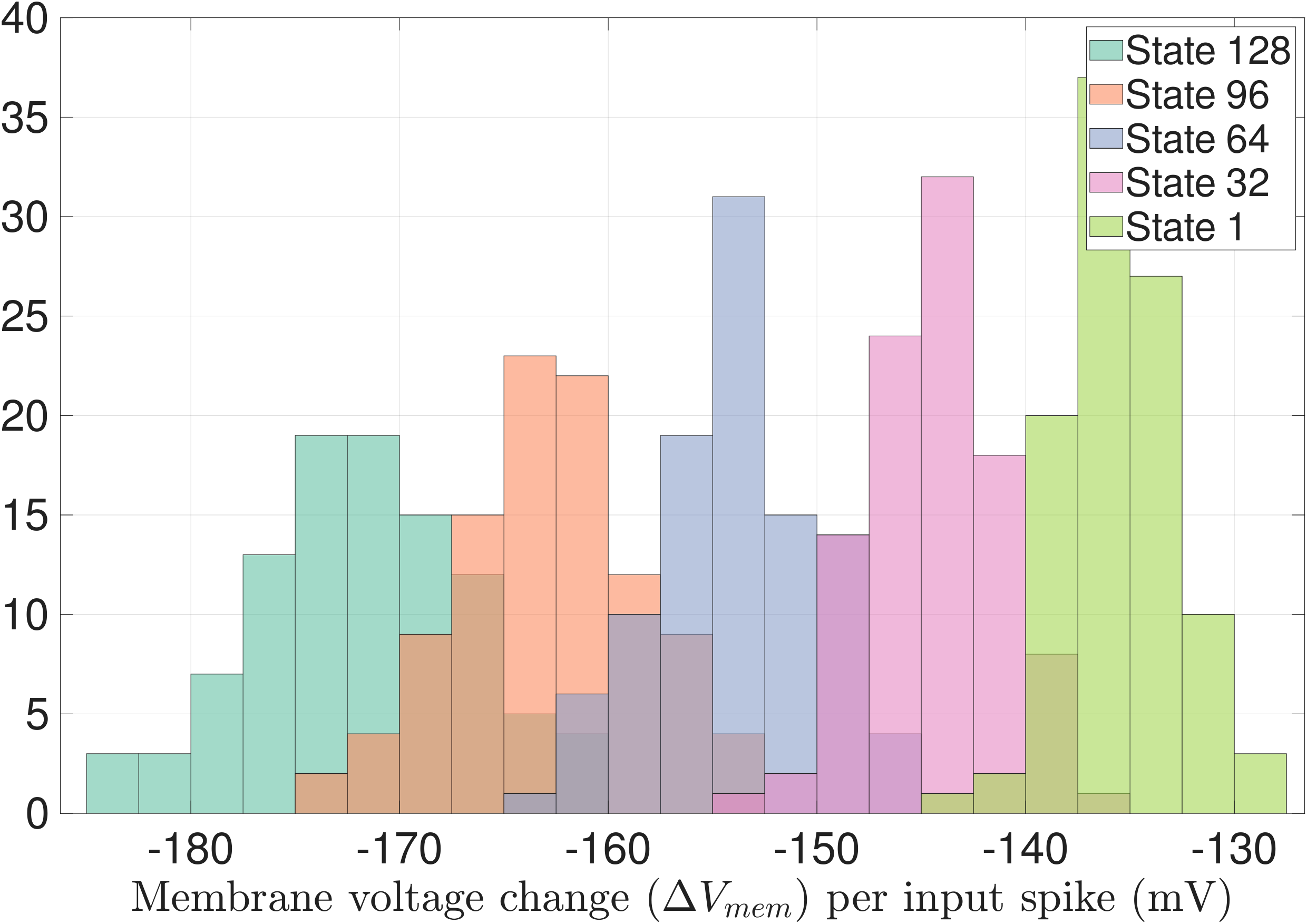}
        (a)
    \end{minipage}
    \begin{minipage}{0.39\textwidth}
        \begin{minipage}{\textwidth}
            \centering
            \includegraphics[width=0.7\textwidth]{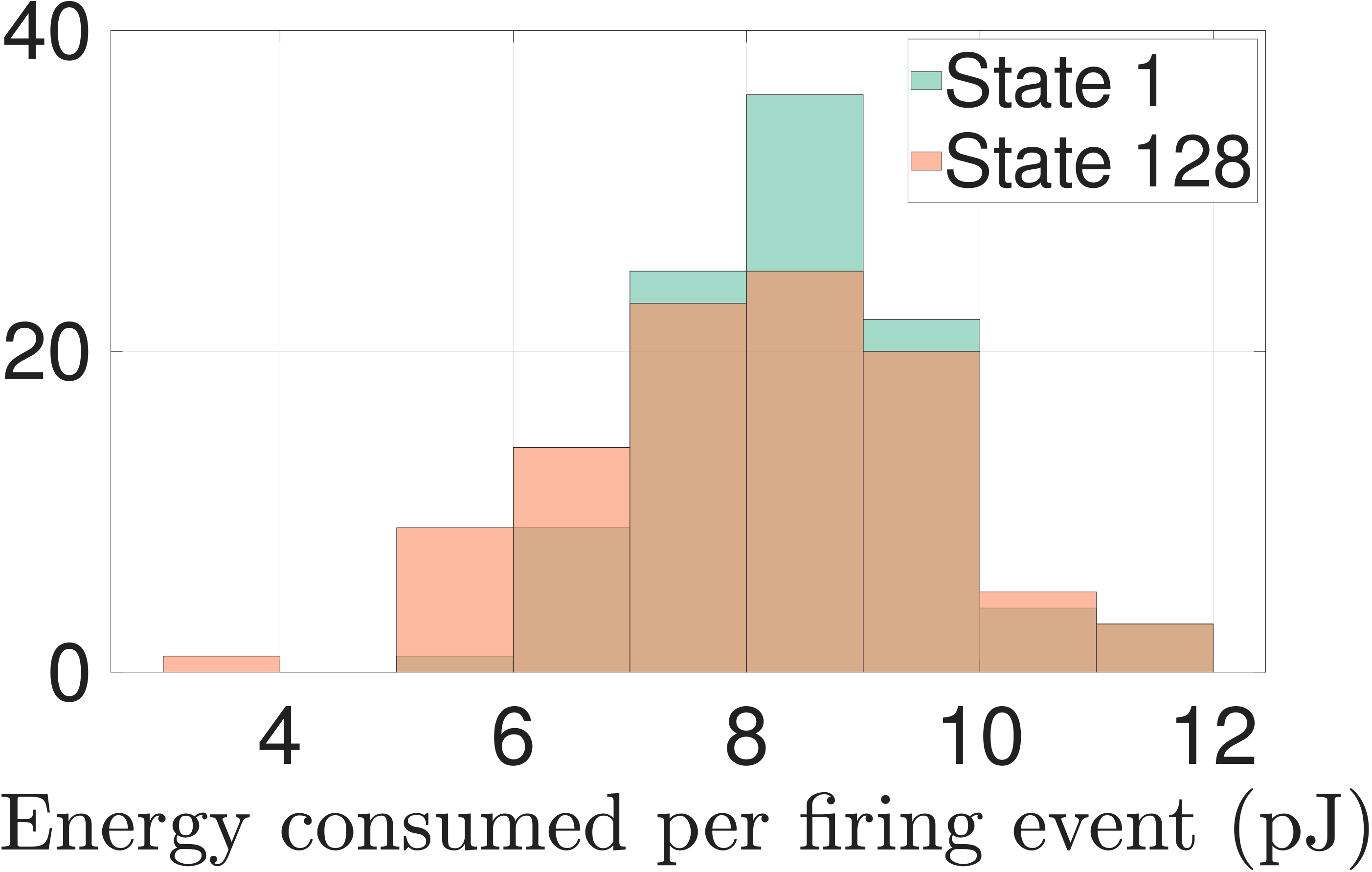}
            
            (b)
        \end{minipage}
        \begin{minipage}{\textwidth}
            \centering
            \includegraphics[width=0.7\textwidth]{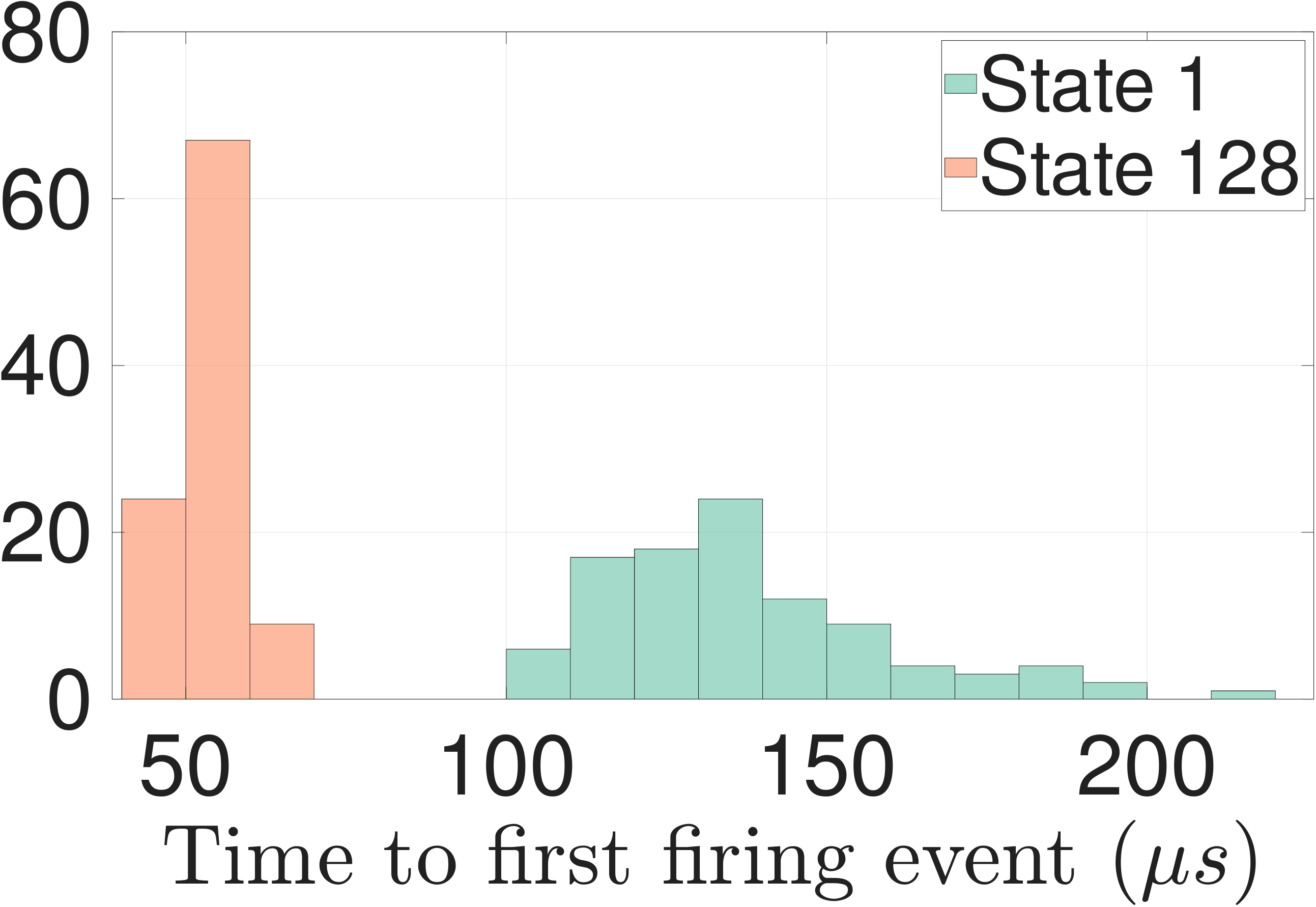}
            
            (c)
        \end{minipage}
    \end{minipage}

    \caption{Histograms demonstrating variability metrics of the delay-feedback LIF topology at selected ECRAM states. (a) The drop in membrane voltage incurred by a single input spike, in mV. (b) The energy consumed by the circuit over the course of a single firing event. (c) The absolute time required to generate an output spike given a 100 kHz, 10\% duty cycle input.}
    \label{fig:histograms}
    \end{mdframed}
\end{figure}

To evaluate robustness against mismatch, we performed a variability analysis through a 100-point Monte-Carlo simulation. For selected ECRAM states 1, 32, 64, 96, and 128, we measured three metrics: the drop in membrane potential for each input AP, the absolute time required for a neuron to reach a firing threshold, and the energy consumed during the firing event (figure \ref{fig:histograms}). 

The drop in membrane potential for each input AP ranges between 100 to 200 mV over the range of ECRAM states, demonstrating the expected modulation of synaptic strength through device conductance. The mean voltage drop associated with each state is approximately evenly spaced, reflecting the strong linearity of the ECRAM conductance states. A larger variance is observed at higher conductance states. This behavior arises because larger synaptic conductance produces greater current draw from the membrane capacitor, making the resulting voltage drop more sensitive to local variations in the membrane capacitance $C_{mem}$.

The energy required to generate a firing event remains consistently low across all Monte Carlo samples and is largely independent of the synaptic weight. Across the full range of ECRAM states, the mean energy consumption is approximately 8 pJ per spike, with a maximum observed value of 12 pJ.

In contrast to the voltage variation trend, the time required to generate an output spike exhibits greater variance at lower conductance states. When synaptic conductance is low, the resulting membrane voltage increments are smaller and more likely to place $V_{mem}$ near the firing threshold. Under these conditions, small variations in the threshold voltage of the output stage produce larger fluctuations in the spike timing.

Despite these variations in membrane voltage and spike timing, the fundamental functionality of the proposed LIF circuit is preserved across all Monte Carlo samples. The observed variability, therefore, primarily affects precise timing characteristics rather than the neuron’s ability to generate spikes. In practice, such variability primarily affects precise spike timing rather than neuron functionality and can be accommodated through circuit calibration and network-level training procedures.

